\documentclass[a4paper,fleqn]{cas-dc}

\usepackage[numbers]{natbib}
\usepackage{tabularx}
\usepackage{bbding}
\usepackage{makecell}
\usepackage{array}
\usepackage{multirow}
\usepackage{hyperref}
\hypersetup{
    colorlinks=true,
    linkcolor=blue,
    urlcolor=blue,
    anchorcolor=blue,
    citecolor=blue
}

\usepackage{float}
\usepackage{tikz}
\usetikzlibrary{backgrounds,mindmap}
\usepackage{pgfplots}
\usepackage{listings}
\usepackage{soul}
\usepackage{color, xcolor}  
\sethlcolor{yellow}

\begin{document}

\shorttitle{Advances in Large Language Models for Medicine} 
\shortauthors{Zhiyu Kan \textit{et al.}}

\title [mode = title]{Advances in Large Language Models for Medicine}   
\author[1]{Zhiyu Kan}
\ead{kjinyi012@gmail.com}
\address[1]{Jinan University, Guangzhou 510632, China}

\author[1]{Wensheng Gan}
\ead{wsgan001@gmail.com}

\author[2]{Zhenlian Qi}
\ead{qzlhit@gmail.com}
\address[2]{Guangdong Eco-Engineering Polytechnic, Guangzhou 510520, China}
\cortext[cor1]{Corresponding author}
\cormark[1]

\author[3]{Philip S. Yu}
\ead{psyu@uic.edu}
\address[3]{University of Illinois Chicago, Chicago 60607, USA}

\begin{abstract}
  Artificial intelligence (AI) technology has advanced rapidly in recent years, with large language models (LLMs) emerging as a significant breakthrough. LLMs are increasingly making an impact across various industries, with the medical field standing out as the most prominent application area. This paper systematically reviews the up-to-date research progress of LLMs in the medical field, providing an in-depth analysis of training techniques for large medical models, their adaptation in healthcare settings, related applications, as well as their strengths and limitations. Furthermore, it innovatively categorizes medical LLMs into three distinct types based on their training methodologies and classifies their evaluation approaches into two categories. Finally, the study proposes solutions to existing challenges and outlines future research directions based on identified issues in the field of medical LLMs. By systematically reviewing previous and advanced research findings, we aim to highlight the necessity of developing medical LLMs, provide a deeper understanding of their current state of development, and offer clear guidance for subsequent research.
\end{abstract}


\maketitle

\section{Introduction}  \label{sec: introduction}

The rapid advancement of AI technology and the breakthroughs in large language models (LLMs) \cite{minaee2024large}, such as the cutting-edge generative pre-trained transformer (GPT) \cite{radford2018improving} series, are transforming the medical industry with unprecedented depth and breadth. With their exceptional performance in text generation, deep understanding, and complex reasoning, these models are driving the medical industry toward greater efficiency and intelligence (model-as-a-service, MaaS \cite{gan2023model}). In the medical industry, the acquisition and processing of information are particularly crucial. Handling the vast amount of medical information is not only essential for supporting medical practitioners in making daily diagnostic and treatment decisions but also indispensable for patients seeking health guidance and researchers exploring the mysteries of diseases. This diagnostic and treatment information spans multiple dimensions, including detailed case data, a rich medical knowledge base, authoritative therapeutic protocols, the latest developments in drug research, disease prevention strategies \cite{liu2025screens}, and findings from health promotion research. The ability to process information directly affects patient diagnosis and treatment, as well as the overall quality of medical services, making it an indispensable pillar in driving medical progress. However, in practice, medical practitioners often find traditional solutions inadequate when faced with the vast amount of medical literature \cite{swanson1990medical} and complex diagnostic case records. 

These capabilities of LLMs inevitably raise questions about whether they will soon replace doctors. To address this issue, we consult ChatGPT, which responds that despite the historic progress made by LLMs in the medical field. They still face three major challenges in their evolution. First, AI technology is still immature. Despite the numerous outstanding features of LLMs, AI technology remains in continuous development and cannot fully replace doctors in applying professional knowledge and skills to address patients' problems. Second, there is the issue of data bias: the quality and accuracy of training data determine the performance of LLMs, and if the data used for training is biased, the capabilities of the resulting LLM will likewise be affected. Finally, there are privacy and security concerns. Medical data is private patient information, and if it is leaked or misused, the consequences can be serious. Therefore, privacy and security must be prioritized in LLM applications.

Generative LLMs refer to deep learning models capable of automatically generating natural language text \cite{peng2023study}. They undergo training using extensive text data and, through a deep understanding of the internal rules of language, automatically produce natural language text that conforms to grammatical and semantic rules. This type of model not only generates coherent and logical text content but also demonstrates strong creativity and generalization capabilities, performing well across different domains and tasks. Meanwhile, there is also a type of discriminative LLM \cite{chow2024unified}, designed to distinguish between different categories or identify patterns in data. These models are commonly used for tasks such as classification, regression, and detection, learning how to differentiate between different outputs or results based on input data. Their differences are outlined in Table \ref{one}. Considering the specific circumstances in the medical field, most medical LLMs are generative, so "LLMs" will be used interchangeably with "generative LLMs" in the following text. The technical principles of LLMs are mainly based on deep learning and natural language processing (NLP). By extensively collecting and training on massive datasets, LLMs can deeply learn and grasp the internal structure and universal patterns of language. These models usually adopt an end-to-end training approach, establishing a mapping relationship between text input and output. Key technologies such as the Transformer model \cite{popel2018training} employ a self-attention framework to enable complex interactions among elements in text sequences. This effectively addresses issues like information loss and gradient vanishing in long text generation, significantly enhancing model performance. Additionally, LLMs combine language modeling with generation algorithms. The former estimates the probability of sentences in a language, while the latter generates specific text based on these probability distributions. Together, they form the core technical framework of generative LLMs.

\begin{table*}[!htp]
    \centering
    \footnotesize
    \caption{Comparison between generative and discriminant formulas}
    \label{one}
    \begin{tabular}{|c|c|m{4cm}<{\centering}|m{4cm}<{\centering}|m{4.5cm}<{\centering}|}
        \hline
         \textbf{Type} & \textbf{Core} & \textbf{Target} & \textbf{Application scenarios} & \textbf{Model complexity} \\
        \hline
        Discriminant & Analysis & Learn the conditional probability distribution of data and directly classify or regress it & Text classification, regression tasks, sequence annotation, etc & Usually simple, only need to model the conditional probability between features and labels\\
        \hline
        Generative & Creation & Learning the joint probability distribution of data can generate new samples & Language modeling, image generation, data augmentation, etc & Usually more complex as it requires modeling the joint distribution of data \\
        \hline
    \end{tabular} 
 \end{table*}

\begin{figure}
    \centering
   \includegraphics[scale=0.31]{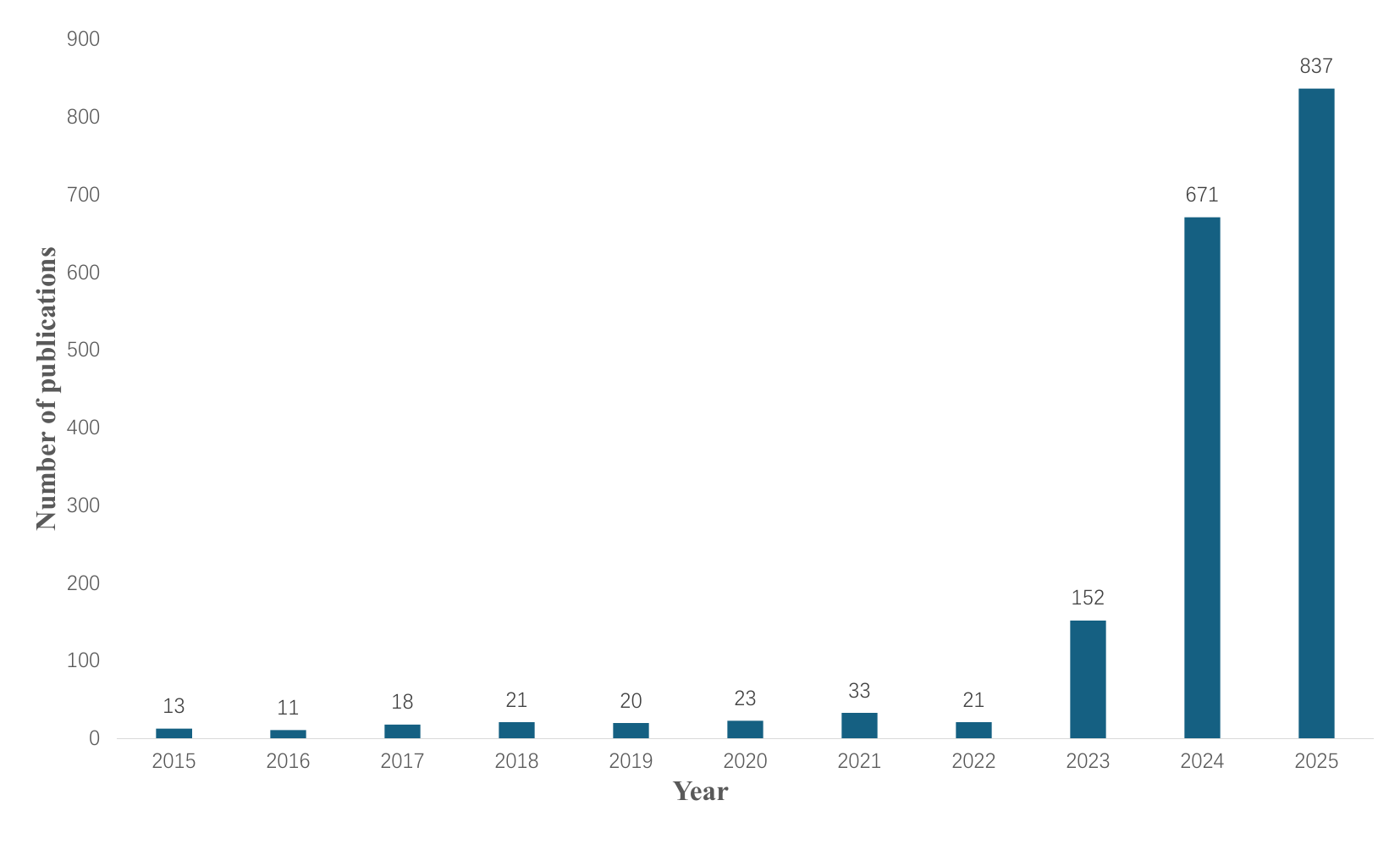}
    \caption{Publication of papers on LLMs in medicine (the data source is derived from the PubMed keyword search for "LLM medicine", with data from 2025 as of September 2025).}
    \label{fig:publi}
\end{figure}

In the fast-paced medical field, LLMs offer transformative potential to enhance clinical practice, medical education, and research \cite{thirunavukarasu2023large}. The early use of LLMs in the medical field was based on a general pre-trained language model for domain adaptation in medical language models. Through continuous technological advancements, LLMs evolve into cutting-edge models such as GPT-4 \cite{achiam2023gpt}. These models overcome the limitations of traditional methods and are capable of handling more complex language processing and understanding tasks. By analyzing medical data and developing more precise diagnoses and personalized treatment plans for doctors and other healthcare professionals, LLMs have the potential to revolutionize the healthcare industry \cite{santamato2024exploring}. This technology not only equips doctors and healthcare workers with unprecedentedly powerful tools but also profoundly reshapes our understanding and practice of disease diagnosis and treatment pathways. Through deep mining and analysis of massive medical data, LLMs can simulate and create new content that is highly similar to or even more accurate than the original information by leveraging advanced deep learning architectures and fine-tuned machine learning models to ensure effective knowledge transmission and innovation. Specifically, the potential transformative impact of LLMs in the healthcare industry is mainly manifested in three areas. First, they provide a new dimension for medical data evaluation, making previously difficult-to-capture subtle changes and trends visible, which offers strong support for early disease detection and intervention. Second, at the diagnostic level, AI algorithms can generate more accurate diagnostic opinions—sometimes surpassing human experience—based on complex data analysis, significantly improving diagnostic accuracy and efficiency \cite{karagounis2024leveraging}. Finally, in treatment planning, LLMs can help ensure that each patient receives personalized treatments based on their individual characteristics, disease progression, and treatment response, truly realizing the vision of precision medicine \cite{pahune2024large}. These LLMs, professionally driven by data and trained through multiple iterations, not only possess strong diagnostic judgment but also respond quickly to doctors' needs in diagnosing rare cases and formulating treatment strategies. They provide valuable references for doctors by retrieving relevant medical literature, case analyses, and expert recommendations \cite{mao2023ad,agbavor2022predicting}. Meanwhile, medical LLMs demonstrate potential capabilities in many fields, including dentistry \cite{huang2023chatgpt}, radiology \cite{akinci2023large}, nuclear medicine \cite{alberts2023large}, and clinical practice \cite{singhal2023large}. Currently, numerous research studies on medical LLMs are in full swing, with a rising trend in the publication and citation of related papers, as shown in Figure \ref{fig:publi}, exhibiting broad application prospects.

Therefore, it is crucial to explore the practical applications, advantages, drawbacks, and potential development pathways of LLMs in the medical field. This latest survey focuses on outlining the extensive application scenarios of large-scale language models in the medical field, exploring in detail the many benefits they bring, the main challenges they currently face, and looking forward to their emerging development trends. By systematically reviewing previous research findings, we aim to explain the necessity of the development of medical LLMs, deeply grasp the current development status of medical LLMs, and provide directional suggestions for subsequent research. To clarify the contributions of this paper and support further research, we conduct a comparative analysis with similar review articles, as shown in Table \ref{two}, highlighting the unique perspective and contributions of our work. Specifically, the contributions of this paper can be summarized as follows:

\begin{itemize}
    \item Comprehensive coverage: This paper provides the latest and most comprehensive review of medical LLMs, covering theoretical foundations and methodological advancements and offering an in-depth examination of their multifaceted medical applications.
    
    \item Progressive review: By reviewing the development history of LLMs, we highlight their characteristics, achievements, and limitations at various stages.
    
    \item Innovative classification: This review innovatively divides medical LLMs into three categories based on training methods and divides their evaluation methods into machine evaluation and human-centered evaluation, providing a new perspective for the research and categorization of medical LLMs.
    
    \item  Trend analysis and strategic suggestions: This paper provides an in-depth analysis of current trends in the field of medical LLMs, including technological advancements and existing challenges, and proposes targeted opportunity identification and future development strategies, aiming to offer effective guidance to researchers and practitioners in related fields.
\end{itemize}

\begin{figure*}
    \centering
    \includegraphics[scale=0.8]{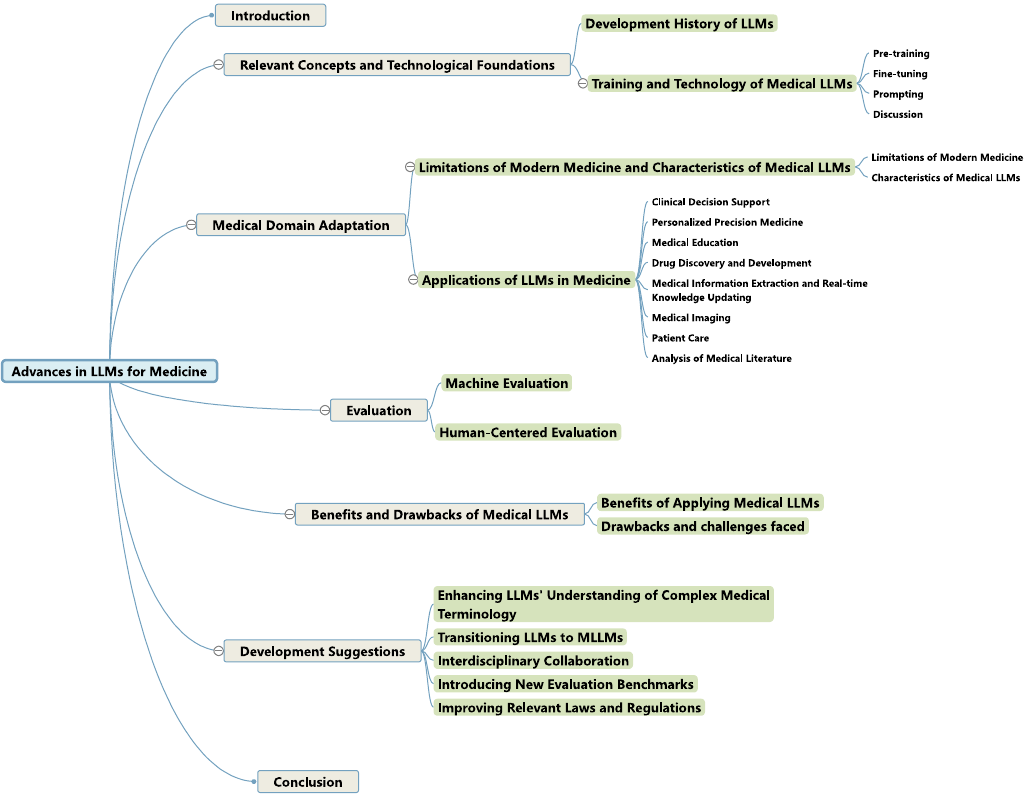}
    \caption{The outline of our overview.}
    \label{fig:outline}
\end{figure*}

\begin{table*}[ht]
    \centering
    \footnotesize
    \caption{Comparison with similar review articles} 
    \label{two}
    \begin{tabular}{|c|c|m{2.5cm}<{\centering}|m{3cm}<{\centering}|c|m{3cm}<{\centering}|c|m{1.2cm}<{\centering}|}
        \hline
          \textbf{Ref.} & \textbf{Year} & \textbf{Summary} & \textbf{Application of LLMs} & \textbf{Comparison} & \textbf{Limits and challenges} & \textbf{Solution} & \textbf{Evaluation}\\
        \hline
        \cite{su2024based} & 2024 & The now and future of LLMs in medicine & Education, scientific research, clinic, nursing	& $\times$ & Humanities, algorithm, and legal ethics	& $\checkmark$ & $\times$ \\
        \hline
        \cite{ullah2024challenges} & 2024 & Challenges and barriers of using LLMs	& Knowledge access, diagnostic support, accessibility, and scalability, continuous learning	& $\times$	& Continuous learning, limited interpretation, ethical and legal concerns, overreliance and dependency & $\checkmark$ & $\times$\\
        \hline
        \cite{bhattacharya2024large} & 2024 & LLMs to multimodal LLM & Drug discovery and development, molecular biology, medical sciences, medical education & $\checkmark$ & Ethical concerns, coherence, accuracy, recency, transparency and interpretability & $\checkmark$ & $\times$ \\
        \hline
        \cite{thirunavukarasu2023large} & 2023	&  LLM technologies in medicine &  On-demand interactive teaching, decision aids, critical appraisal& $\checkmark$ & Timeliness, exactitude, coherence, transparency and interpretability, ethics & $\checkmark$ & $\times$\\
        \hline
        \cite{waisberg2024concerns} & 2024	& Concerns with LLMs in medicine & $\times$ & $\times$ & Accuracy, ethical concern, technology outflow	& $\times$ & $\times$\\
        \hline
        \cite{shool2025systematic} & 2025 & A review of LLM evaluations in clinical medicine & $\times$ & $\checkmark$ & Technical performance defects, clinical application barriers, ethical and social risks & $\checkmark$ & $\checkmark$ \\
        \hline
        \cite{zheng2025large} & 2025 & LLMs for medicine: a survey	& Medical education, guideline, medical robotics, translation & $\times$ & Hallucination, domain data limitations, lack of evaluation benchmarks and metrics, ethical and safety concerns & $\checkmark$ & $\times$ \\
        \hline
        Our work & 2025	& Advances in LLMs for medicine & 	Assistant decision, individualized targeted therapy, education, drug design, medical imaging	& $\checkmark$	& Confusing memories, troublesome adding and deleting, lack of unified evaluation, humanistic ethics	& $\checkmark$ & $\checkmark$\\
        \hline       
    \end{tabular}
 \end{table*}

Literature review methodology: To ensure a comprehensive and professional investigation of LLMs in the medical field, we adopt a systematic review approach. Specifically, we conduct searches in major academic databases such as Web of Science, DBLP, IEEE Xplore, and Google Scholar using "LLM medicine" and "medical LLM" as core keywords. While prioritizing publications from the 2020-2025 period, we also included selected foundational studies published before 2020 that significantly advanced the field. Initially, the retrieved literature was screened for relevance based on titles and abstracts. After the initial screening of titles and abstracts, candidate literature was further refined according to predefined steps to ensure the selected studies were both relevant and of high quality. 1. Inclusion criteria: Reviews proposing innovative methodologies, theoretical analyses, or experimental validation of their own viewpoints; literature examining the process from training to deployment of one or more medical LLMs; formally peer-reviewed English publications. 2. Exclusion criteria: Articles that merely compile others' research findings without providing original insights; publications without formal peer review; studies that match the keywords but have low direct relevance to LLMs in the medical field.

As shown in Figure \ref{fig:outline}, this review seeks to answer the following questions. Section \ref{sec:relatedwork}: How have LLMs developed? How are medical LLMs trained? Section \ref{Medical Domain Adaptation}: What are the limitations of modern medicine? What capabilities do medical LLMs offer beyond modern medicine? Section \ref{evaluation}: How are the current medical LLMs evaluated? Section \ref{BenefitsandDrawbacks}: What are the advantages and disadvantages of applying medical LLMs? Section \ref{sec:directions}: What is the future direction for the development of medical LLMs? Section \ref{sec:Conclusion} makes a conclusion.

\begin{table}[ht]
    \footnotesize
    \caption{Abbreviations of commonly used terms throughout this review and their complete forms} 
    \begin{tabular}{|c|m{5cm}<{\centering}|}
        \hline
        \textbf{Abbreviation} & \textbf{Complete form} \\
        \hline
        BERT & Bidirectional Encoder Representations from Transformer \\
        \hline
        CoT & Chain-of-Thought \\
        \hline
        EHR & Electronic Health Record \\
        \hline
        FTORs & Free-Text Oncology Reports \\
        \hline
        GPT & Generative Pre-trained Transformer\\
        \hline
        ICL & In-Context Learning \\
        \hline
        IFT & Instruction Fine-Tuning \\
        \hline
        
        LLMs	& Large Language Models \\
        \hline
        MCQA & Multiple-choice QA \\
        \hline
        MLLM & Multimodal Large Language Model \\
        \hline
        MLM &  Masked Language Modeling \\
        \hline
        MRI & Magnetic Resonance Imaging \\
        \hline
        MTQA & Multiple-turn QA \\
        \hline
        NLP	& Natural Language Processing \\
        \hline
        
        NSP & Next Sentence Prediction \\
        \hline
        NTP & Next Token Prediction \\
        \hline
        PaLM & Prompt Language Model \\
        \hline
        PEFT & Parameter-Efficient Fine-Tuning \\
        \hline
        PLM	& Protein Language Model \\
        \hline
        RAG	& Retrieval-Augmented Generation \\
        \hline
        
        SFT & Supervised Fine-Tuning \\
        \hline
        STQA &  Single-turn QA \\
        \hline
        TRCs & Tumor Response Categories \\
        \hline
    \end{tabular}
    \label{three}
 \end{table}

\section{Relevant Concepts and Technological Foundations} \label{sec:relatedwork}
\subsection{Development History of LLMs}

The development journey of LLMs can be divided into six stages. In 1948, N-gram models appeared \cite{majumder2002n}. It could predict the probability of the next word by statistically analyzing the frequency of consecutive n-words in text, thereby establishing language characteristic models. However, these models had limitations in capturing complex language dependency relations and deep semantic structural information. Subsequently, the bag-of-words model \cite{zhang2010understanding} came out in 1954, considering the text as a group of unordered words for text feature extraction, but ignoring the order of words before and after, and contextual semantic relationships. By 2003, neural probabilistic language models \cite{bengio2003neural} incorporated neural networks, aiming to enhance model performance through neural network learning capabilities. Due to the limitations of neural network size and training datasets at that time, the processing capabilities of these models were still suboptimal. Then, in 2013, the Word2Vec model \cite{jatnika2019word2vec} was introduced. It captured semantic relationships between words by training and learning word vectors, opening a new avenue for improving the performance of LLMs. A pivotal breakthrough occurred in 2018 with the introduction of pre-training LLMs such as BERT \cite{devlin2019bert} and GPT \cite{radford2018improving}. Built upon the Transformer architecture, these models were pre-trained on extensive text data, acquiring a profound understanding of both semantics and syntax. The year 2020 witnessed the release of the T5 model \cite{raffel2020exploring}, which reframed diverse NLP tasks as text-to-text transformation challenges, achieving remarkable results. Moreover, within a span of 2 to 3 years, various mixture-of-experts \cite{gan2025mixture} (MoE)-based models, such as GPT-4 \cite{achiam2023gpt} and GPT-3.5 \cite{bhaskar2023prompted}, underwent further optimization and improvement in terms of their scale, function, and generation capabilities. So far, LLMs have been successfully applied to many different fields, including education \cite{gan2023large}, law \cite{lai2024large}, medicine \cite{zheng2025large}, robotics \cite{zeng2023large}, etc.

\subsection{Training and Technology of Medical LLMs}

When LLMs transition from general-purpose to medical applications, it is akin to transitioning from toys to tools, where precision in the tool is of utmost importance. The training process for LLMs in the medical field primarily consists of three key steps: pre-training based on textual data \cite{zeng2023distributed,zeng2025distributed}, fine-tuning based on QA datasets \cite{hu2023llm}, and prompt engineering based on task instructions \cite{liu2023pre}, to adapt general models to the specific needs of the medical field.

\subsubsection{Pre-training}

In the medical field, pre-training typically refers to utilizing large-scale medical corpora (such as electronic health record (EHR) data \cite{peng2023study}, clinical guidelines, practice norms, medical textbooks, academic papers, etc.) as training resources. Through self-supervised learning, the model learns medical terminology and the deep semantic associations between these terms and complex language structures without explicit labels. Medical LLMs undergo specialized pre-training by refining three fundamental learning paradigms derived from general-domain architectures: masked language modeling (MLM) \cite{devlin2019bert}, next sentence prediction (NSP) \cite{devlin2019bert}, and next token prediction (NTP) \cite{tom2020brown}. The implementation varies by architectural paradigm. Encoder-style systems such as BioBERT \cite{lee2020biobert} and PubMedBERT \cite{gu2021domain} leverage both MLM \cite{devlin2019bert} and NSP  \cite{devlin2019bert} synergistically, whereas decoder-focused frameworks, including BioGPT \cite{luo2022biogpt}, exclusively employ NTP \cite{tom2020brown} optimization. Notably, the medical BERT-family architectures represent domain-specialized adaptations built upon foundational BERT/RoBERTa infrastructures through continued medical corpus training. After this pre-training phase, the model goes through a fine-tuning procedure on smaller and specialized datasets to improve its performance and accuracy for specific medical tasks. This process equips the model with a profound understanding of the characteristics of medical language, from grammatical structures to semantic connotations, enabling it to effectively grasp medical professional vocabulary, terminology, and their applications in specific contexts. Meanwhile, it accurately captures the complex sentence structures and conceptual relationships in medical language, thereby constructing a solid medical language knowledge system and acquiring exceptional medical language understanding capabilities. For example, the Biomistral \cite{labrak2024biomistral} model significantly improves its accuracy and efficiency in medical question-answering and multilingual complex evaluation tasks through pre-training on the PubMed literature dataset. The clinical BERT \cite{alsentzer2019publicly} model, after pre-training on a vast EHR dataset, significantly enhances its automated processing capabilities for EHR data, demonstrating remarkable performance in medical application scenarios such as patient history summarization, clinical event extraction, and medical record parsing.

\subsubsection{Fine-tuning}

Developing medical-specific LLMs through full pretraining incurs substantial costs in both computational power (often requiring weeks of processing time) and human resources. As an alternative solution, scientists devise domain adaptation strategies that leverage pretrained general-purpose models through specialized tuning techniques \cite{singhal2025toward,toma2023clinical}. Three primary approaches emerge: Supervised Fine-Tuning (SFT)  \cite{ouyang2022training} for continuous learning from medical text corpora, Instruction Fine-Tuning (IFT) \cite{jason2022finetuned} for processing annotated clinical directives, and Parameter-Efficient Fine-Tuning (PEFT) \cite{houlsby2019parameter} to minimize hardware requirements during model customization. The SFT approach essentially continues the pretraining process using medical domain data, while IFT \cite{jason2022finetuned} specializes in handling structured clinical instructions. The PEFT \cite{houlsby2019parameter}  methodology particularly addresses the challenge of excessive resource consumption during model adaptation phases.

\textbf{Supervised Fine-Tuning (SFT)} \cite{ouyang2022training} facilitates medical domain adaptation of general language models through integration of high-quality healthcare datasets comprising physician-patient interactions \cite{li2023chatdoctor}, clinical question-answering records \cite{han2023medalpaca}, and structured medical knowledge representations \cite{ye2023qilin}. This process extends conventional pretraining by maintaining identical learning objectives (e.g., token prediction) while focusing on biomedical content. The technique effectively establishes a transitional phase between general language understanding and medical specialization, enabling models to develop healthcare-specific competencies through targeted knowledge acquisition and conceptual alignment. Various medical LLM variants have emerged through application-specific dataset selection: Dialogue-optimized adaptations like DoctorGLM \cite{xiong2023doctorglm} (derived from ChatGLM \cite{du2022glm}) and ChatDoctor \cite{li2023chatdoctor} (based on LLaMA \cite{touvron2023llama}) utilize clinical conversation transcripts, while QA-enhanced versions such as MedAlpaca \cite{han2023medalpaca} (built upon Alpaca \cite{taori2023stanford}) incorporate extensive medical question-answering pairs. More comprehensive approaches like Clinicalcamel \cite{toma2023clinical} combine multiple data types (dialogues, literature, and QAs) to refine LLaMA-2 \cite{touvron2023llama}, whereas knowledge-infused systems, including Qilin-Med \cite{ye2023qilin} and Zhongjing \cite{yang2024zhongjing} leverage graph-based medical ontologies to augment Baichuan \cite{yang2023baichuan} and LLaMA \cite{touvron2023llama} architectures, respectively. Extensive validation confirms this refinement approach's capacity to simultaneously enhance both biomedical language processing proficiency and clinical decision-support reliability \cite{he2025survey}, thereby offering a robust framework for developing healthcare-specific variants of general-purpose language models.

\textbf{Instruction Fine-Tuning (IFT)} \cite{jason2022finetuned} develops specialized medical LLMs by creating structured training datasets \cite{he2025survey} containing instruction-input-output sequences (e.g., clinical question-answer pairs), enhancing both instruction compliance and medical domain alignment. This approach contrasts with SFT \cite{ouyang2022training}: while SFT \cite{ouyang2022training} extends pre-training with medical corpora to improve text comprehension and token prediction, IFT \cite{jason2022finetuned} specifically optimizes a model's ability to interpret and respond to clinical instructions \cite{zhang2024gpt4roi}. This distinction leads to divergent data requirements — SFT \cite{ouyang2022training} prioritizes large-scale medical texts, whereas IFT \cite{jason2022finetuned} emphasizes high-quality annotated examples. Recognizing their complementary strengths, current research \cite{zhang2023alpacare} explores hybrid SFT \cite{ouyang2022training} + IFT \cite{jason2022finetuned} strategies for more capable medical LLMs. Effective IFT \cite{jason2022finetuned} implementation demands carefully constructed instructional datasets representing diverse medical contexts. Representative approaches include: BenTsao's \cite{wang2023huatuo} knowledge-graph-derived training materials, Zhongjing's \cite{yang2024zhongjing} conversational dialogue-based curriculum, and MedAlpaca's \cite{han2023medalpaca} combined use of clinical dialogues and QA pairs for instruction tuning.

\textbf{Parameter-Efficient Fine-Tuning (PEFT) methodologies} \cite{houlsby2019parameter} optimize the adaptation of pre-trained LLMs by selectively updating only a minimal subset of parameters (or introducing limited additional parameters) while preserving the majority of the original model's parameters intact. This approach dramatically reduces both computational overhead and memory footprint during the fine-tuning process. The most widely adopted PEFT techniques \cite{houlsby2019parameter} encompass Low-Rank Adaptation (LoRA) \cite{hulora}, Prefix Tuning \cite{li2021prefix}, and Adapter-based refinement \cite{liu2022p}. LoRA functions by incorporating learnable low-rank factorized matrices within the self-attention modules of each Transformer block, while keeping the original model's weights entirely unchanged. This strategy achieves significant parameter reduction while preserving the model's capacity to capture task-specific patterns. Prefix tuning employs task-specific continuous vector embeddings ("prefixes") prepended to each layer's input sequence, providing contextual guidance for generation without modifying core parameters. The Adapter approach \cite{houlsby2019parameter} incorporates compact neural network subunits between Transformer layers, enabling targeted adaptation through these small, trainable modules while keeping the principal model weights fixed. These parameter-efficient strategies demonstrate notable efficacy in clinical domain adaptation, as evidenced by prominent medical LLMs such as Baize-Healthcare \cite{xu2023baize}, CPLLM \cite{ben2024cpllm}, and Clinical Camel \cite{toma2023clinical}, all of which leverage LoRA for efficient domain alignment. The success of these implementations underscores PEFT's \cite{houlsby2019parameter}  critical advantage in maintaining model performance while optimizing resource utilization.

\subsubsection{Prompting }

While fine-tuning significantly lowers computational expenses relative to pre-training, it still involves additional model training and the need for high-quality fine-tuning datasets, consuming both computational resources and manual effort. In contrast, prompting techniques effectively adapt general-purpose LLMs (e.g., PaLM \cite{chowdhery2023palm}) to medical fields (e.g., Med-PaLM \cite{singhal2023large}) without modifying any model parameters. Widely used prompting approaches include In-Context Learning (ICL) \cite{tom2020brown}, Chain-of-Thought (CoT) prompting \cite{wei2022chain}, prompt tuning \cite{lester2021power}, and Retrieval-Augmented Generation (RAG) \cite{lewis2020retrieval}.

\textbf{In-Context Learning (ICL)} \cite{tom2020brown} is an approach that enables a model to generate desired outputs by incorporating specific prompts or contextual information directly into the input, without the need for further training of model parameters. The two most widely used forms of ICL are zero-shot and few-shot prompting. Zero-shot prompts provide explicit instructions without including concrete examples, thereby guiding the LLM to effectively accomplish the intended task based on the given directives \cite{tang2024instance}. In contrast, few-shot prompts supply a limited number of demonstrations or illustrative examples prior to task execution, assisting the model in recognizing underlying patterns and leveraging its inherent knowledge and reasoning capacity to produce task-appropriate responses \cite{yuan2024generalized}. A notable application of this technique is Med-PaLM \cite{singhal2023large}, which is built upon the PaLM architecture \cite{chowdhery2023palm}. By training on a small set of medical examples, it achieves accuracy in responding to both multiple-choice and open-ended medical questions that rivals the performance of human clinicians.

\textbf{Chain-of-Thought (CoT) Prompting} \cite{wei2022chain} enhances model performance beyond standard in-context learning by improving both accuracy and logical coherence. This approach guides models to articulate intermediate reasoning steps when solving complex problems through carefully designed prompt phrasing \cite{wei2022chain}. When integrated with few-shot learning, CoT \cite{wei2022chain} enables medical LLMs to demonstrate their reasoning process while formulating responses. Empirical studies confirm CoT's \cite{wei2022chain} effectiveness in boosting performance on complex reasoning tasks like medical question answering \cite{singhal2023large}. Leading medical LLMs—including DeID-GPT \cite{liu2023deid}, Med-PaLM \cite{singhal2023large}, and MedPrompt—leverage \cite{nori2023can} this technique to replicate clinical diagnostic reasoning, delivering more transparent and interpretable outputs. Notably, MedPrompt achieves state-of-the-art results on medical QA tasks using GPT-4 \cite{achiam2023gpt} with CoT prompting \cite{wei2022chain} alone, surpassing specialized fine-tuned medical models without any parameter updates—demonstrating the power of advanced prompting strategies in healthcare AI applications.

\textbf{Prompt Tuning} \cite{lester2021power} enhances model capabilities by integrating both prompting and fine-tuning approaches \cite{liu2022p}. This technique employs trainable continuous vectors as learnable prompts, which can be adaptively optimized during the fine-tuning phase to suit diverse medical applications and objectives. Unlike conventional discrete prompting methods that rely on fixed templates, prompt tuning \cite{lester2021power} offers greater flexibility in guiding LLMs. While traditional fine-tuning requires updating all model parameters, prompt tuning \cite{lester2021power} selectively optimizes only a minimal set of prompt-related parameters, significantly reducing computational overhead. This approach provides cost-effective yet precise solutions for medical applications \cite{nori2023can}, delivering high performance with significantly lower training costs compared to full-model fine-tuning.

\textbf{Retrieval-Augmented Generation (RAG)} \cite{lewis2020retrieval} improves LLM outputs by dynamically incorporating external knowledge sources. This approach addresses three key LLM limitations: hallucination generation, opaque reasoning patterns, and dependence on obsolete data through real-time knowledge base integration \cite{gao2023retrieval}. The framework operates through three sequential phases: information retrieval from external sources, contextual augmentation of prompts, and knowledge-guided response generation. Unlike conventional fine-tuning, which modifies model parameters, RAG \cite{lewis2020retrieval} preserves existing knowledge while seamlessly incorporating new information through database updates. This characteristic makes it particularly valuable for high-stakes, fast-evolving domains such as healthcare. The recently developed MIRAGE benchmark \cite{xiong2024benchmarking}, comprising 7,663 medical questions from five QA datasets, provides standardized evaluation metrics for advancing medical RAG \cite{lewis2020retrieval} applications. The retrieval mechanism relies on semantic similarity computations between query and document embeddings, with model selection critically impacting performance. Current leading embedding systems include AngIE \cite{li2023angle}, Voyage \cite{wang2023voyager}, and BGE \cite{xiao2024c}. Advanced retrieval optimization techniques encompass adaptive query refinement \cite{shao2023enhancing}, multi-level recursive searching \cite{trivedi2023interleaving}, and iterative retrieval improvement \cite{asai2024self}.

\subsubsection{Discussion}

This section categorizes medical LLMs into three types based on their technical approaches: pre-training, fine-tuning, and prompt engineering.  Different implementation strategies can be adopted based on available computing resources. Organizations with powerful computing infrastructure have two viable options: developing medical-specific LLMs through complete pre-training or adapting existing open-source general models like LLaMA \cite{touvron2023llama} via extensive medical data fine-tuning. Research findings from models such as MedPaLM-2 \cite{singhal2023large}, MedAlpaca \cite{han2023medalpaca}, and Clinical Camel \cite{toma2023clinical} confirm that domain-specific fine-tuning of general models leads to substantial improvements in medical task performance \cite{wu2024clinical}. Particularly noteworthy is Clinical Camel \cite{toma2023clinical}, a LLaMA-2-70B \cite{genai2023llama} based fine-tuned model that demonstrates superior capabilities compared to GPT-4 \cite{achiam2023gpt}. For resource-constrained users, including small businesses and individual researchers, alternative effective approaches exist. These involve optimizing commercial black-box LLMs through intelligent integration of medical domain knowledge with techniques like in-context learning, advanced prompting methods, and retrieval-augmented generation. The MedPrompt \cite{nori2023can} case illustrates this approach, where sophisticated prompt engineering enabled GPT-4 \cite{achiam2023gpt} to match or exceed the performance of both specialized fine-tuned models (e.g., MedPaLM-2 \cite{singhal2023large}) and human medical experts. These results indicate that well-designed prompt strategies may offer a computationally efficient and sustainable alternative to traditional fine-tuning approaches in healthcare applications.

\section{Medical Domain Adaptation}
\label{Medical Domain Adaptation}

\subsection{Limitations of Modern Medicine and Characteristics of Medical LLMs}

Medicine, as the art and science of healing \cite{shore2009art}, maintaining, and enhancing human health, as well as preventing and treating diseases, possesses unique properties and characteristics. These properties and characteristics not only determine the complexity and diversity of medicine but also pose specific requirements on the design and application of medical LLMs. In this section, we discuss the characteristics of LLMs, the limitations of modern medicine, and the integration between LLMs and medicine.

\subsubsection{Limitations of Modern Medicine}

\textbf{Specialization-induced limitations} \cite{ackerknecht2016short}. With advancements in technology and increasing specialization, modern medicine has gradually divided into multiple specialized fields. Although this specialization facilitates in-depth research into specific diseases, it may lead to a lack of a holistic perspective among doctors. Each doctor excels only in their area of expertise and has limited knowledge of other medical disciplines. When facing complex cases, consultations among doctors from multiple departments may be required, increasing the complexity and time cost of diagnosis. Furthermore, consultations among multiple departments may bring communication barriers and coordination difficulties. Doctors from different departments may have varying diagnostic concepts and treatment methods, requiring thorough communication and negotiation during the consultation process to ensure consistency and effectiveness of the treatment plan. This process is not only time-consuming and labor-intensive but may also lead to delays or changes in the treatment plan due to poor communication or disagreements.

\textbf{Lack of knowledge and experience} \cite{castiglioni2019history}. Modern medicine is a discipline with a huge amount of information that is constantly changing, requiring doctors to not only have a solid theoretical foundation but also rich practical experience. From basic biology, anatomy, and physiology, to complex pathology, pharmacology, and clinical diagnostic methods, the knowledge system that doctors need to master is vast and complex. However, in some cases, this knowledge and experience may be insufficient to address all conditions, especially chronic and rare diseases. When facing these diseases, doctors often feel powerless and struggle to provide perfect solutions.

\textbf{Challenges of personalized treatment} \cite{lammert2024expert}. The recipients of modern medical services are mostly individuals, each with different physical conditions, disease types, and severity levels, requiring doctors to provide personalized treatment. While doctors’ knowledge and experience can offer general treatment recommendations, implementing care at a detailed level requires strong individual expertise. Therefore, despite recent developments in precision medicine, achieving true personalized treatment for all patients solely relying on doctors' individual capabilities will take time.

\subsubsection{Characteristics of Medical LLMs}

\textbf{Large scale.} In the context of LLMs, the term "large scale" can be understood from two dimensions. On the one hand, the number of parameters in LLMs is extremely large and has shown explosive growth in recent years, rapidly climbing from billions to trillions. For example, Google's BERT in 2018 had 300 million parameters, in 2019 GPT-2 boasted 1.5 billion parameters \cite{radford2019language}, while GPT-3 \cite{brown2020language} in 2021, soared to an impressive 175 billion parameters \cite{brown2020language}. By 2022, the Switch Transformer's parameter count was an astonishing 1.6 trillion \cite{lin2021m6}. On the other hand, the training data for LLMs is equally vast, encompassing sources such as webpages, academic papers, and dialogues. This large-scale dataset allows the models to study and represent complex grammar and relationships in language, thereby excelling in various NLP tasks.

\textbf{High specialization} \cite{ling2023domain}. Medical LLMs focus on the healthcare field, dedicating themselves to the in-depth, meticulous, and comprehensive study of medical knowledge and experience. This high degree of specialization enables medical LLMs to thoroughly understand the complexity and diversity of the medical field. They not only master a wide range of medical theoretical knowledge but also extract valuable experience from extensive clinical practices. Through ongoing learning and refinement, medical LLMs can accurately identify disease symptoms, analyze pathological mechanisms, and offer customized treatment protocols for individual patients. Simultaneously, they can assist doctors in key decisions such as disease diagnosis, surgical planning, and medication selection, advancing the exactitude and productivity of medical services.

\textbf{Extensive application scenarios} \cite{bedi2024testing}.  As specialized models in the medical field, medical LLMs have application scenarios that span nearly all aspects of medicine. For example, clinical decision support \cite{sutton2020overview}, individualized targeted therapy, medical education, and drug development. Additionally, in processing medical literature, they can use NLP algorithms to understand, analyze, and generate natural language \cite{hovy1987generating} text. They can not only automatically extract key information from medical literature, such as research methods, experimental results, and conclusions, providing researchers with convenient literature review tools, but also automatically generate structured medical records based on patients' medical data, boosting productivity and excellence in medical service delivery.

\textbf{High complexity and investment costs.} LLMs are progressively becoming more complex, with single-step computation time increasing by over tenfold. For high-traffic businesses, training experiments that previously took only a few hours now extend to days, while completing tests within 24 hours has become a standard requirement \cite{aryan2023costly}. Moreover, the cost of training a general LLM is substantial. When factoring in subsequent optimization, updates, and deployment, the overall expense escalates further. Take ChatGPT as an example; the estimated cost of training OpenAI's GPT-4 \cite{achiam2023gpt} model exceeds 100 million dollars \cite{xia2024understanding}. Additionally, ChatGPT demands a significant number of GPU chips for data processing. Medical LLMs, being specialized and trained on the foundation of general LLMs, naturally incur even greater costs. For instance, Google's training of the 540-billion-parameter PaLM model consumed vast computing resources, equating to a training cost of roughly twenty-five to thirty million dollars  \cite{chowdhery2023palm}.

\subsection{Applications of LLMs in Medicine}

\begin{figure}
    \centering
    \includegraphics[scale=0.42]{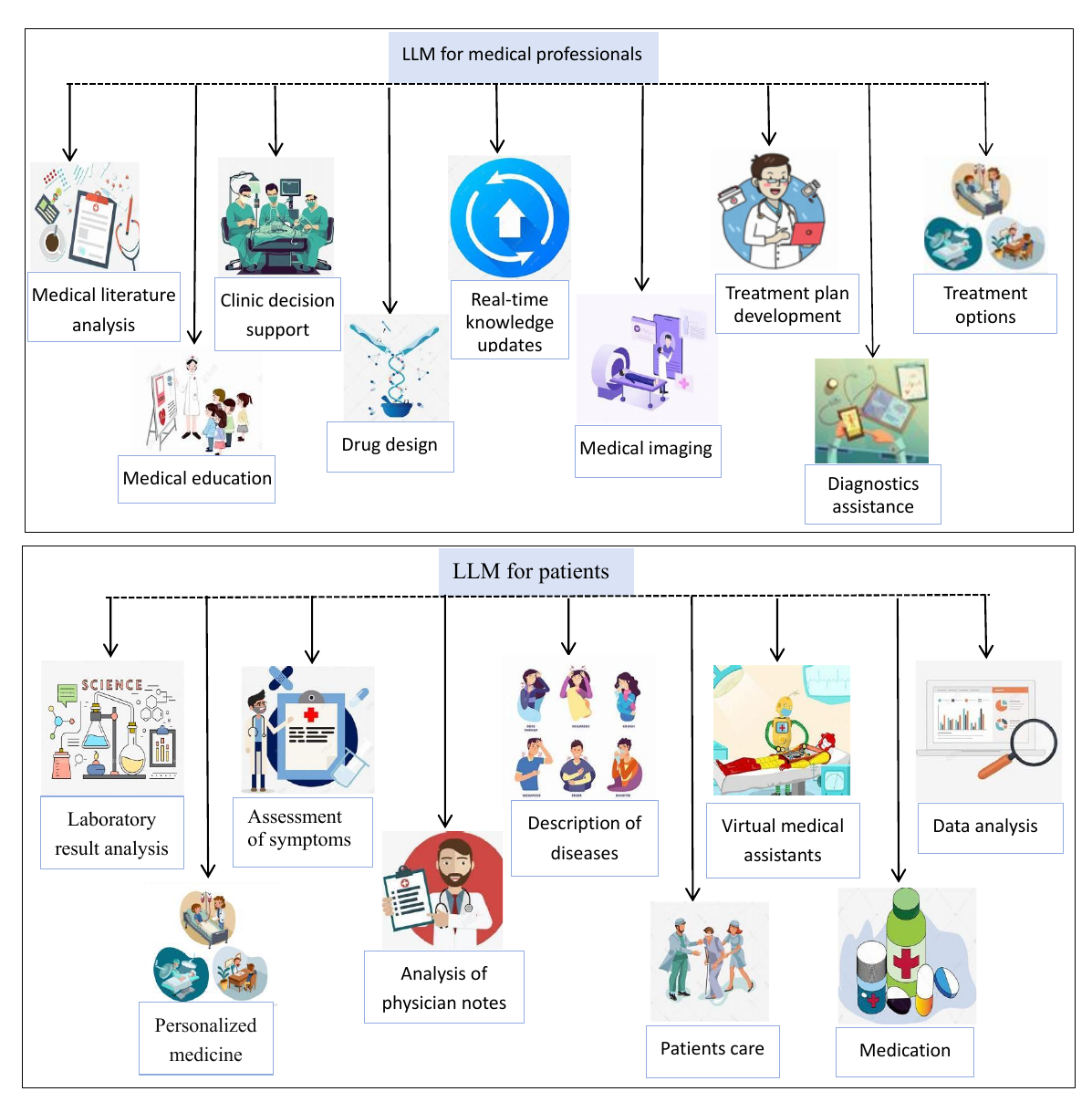}
    \caption{Applications of the medical LLMs.}
    \label{fig:allms}
\end{figure}

LLMs have numerous applications in healthcare, as illustrated in Figure \ref{fig:allms}. Below are the most significant ones.

\subsubsection{Clinical Decision Support}

In the field of medical diagnosis, a series of complex clinical operations are often rooted in in-depth data analysis \cite{lin2015rwfim,gan2023anomaly}, rigorous clinical research, and professional recommendations \cite{chen2023language}. In the medical domain, LLMs exhibit a key application value—serving as powerful tools for clinical decision support. They can provide highly personalized diagnostic insights and treatment recommendations based on each patient's unique condition and physical status, effectively assisting doctors in making more precise and evidence-based decisions in complex and ever-changing clinical environments, thereby optimizing the medical processes and enhancing treatment outcomes. The applications of LLMs extend even further. They are broadening the boundaries of clinical decision-making, optimizing recruitment processes for clinical trials, facilitating efficient management of clinical data, providing strong support for medical research, and playing an active role in patient education \cite{xue2023potential,chen2023boosting}. By utilizing NLP algorithms, generative LLMs can understand, analyze, and generate natural language text. They can not only automatically extract key information from medical literature—such as research methods, experimental results, and conclusions—providing convenient literature review tools for researchers, but also automatically generate structured medical records from patients' medical data, enhancing productivity and excellence in medical service delivery. Notably, some cutting-edge research has attempted to use advanced transformer models such as BERT \cite{devlin2019bert}, RoBERTa, and DistilBERT \cite{adoma2020comparative} to predict COVID-19 diagnosis results by analyzing textual descriptions of acute chemosensory alterations \cite{malla2021covid}. This practice not only verifies the feasibility of generative models in infectious disease diagnosis but also opens the door for exploration in more disease areas \cite{li2023text}. Furthermore, for the diagnosis of complex neurological diseases such as Alzheimer's disease \cite{mao2023ad} and dementia \cite{agbavor2022predicting}, scholars have also explored new pathways using LLMs for auxiliary diagnosis, further enriching the application scenarios of LLMs in the medical field. Additionally, there has been literature advocating the integration of LLM chatbots to meet similar needs \cite{bill2023fine,bilal2024enhancing}.

\subsubsection{Personalized Precision Medicine}

LLMs can serve as virtual health assistants for patients, providing tailored health advice based on patients' medical history, current symptoms, and other related information. LLMs have emerged as disruptive tools, significantly enhancing capabilities in patient care \cite{javaid2023chatgpt}. For example, for patients with common colds or flu, these virtual assistants can intelligently recommend suitable over-the-counter medications or home remedies. Patients can easily access this convenient service through diverse platforms such as websites, mobile applications, and even voice assistants. For patients residing in remote or medically underserved areas, this feature is particularly important as it greatly promotes the accessibility of medical services, enabling patients to obtain professional and personalized medical guidance without traveling long distances to hospitals. By providing personalized advice \cite{eapen2023personalization}, patient-specific treatment protocols coupled with ongoing surveillance of patients' progress throughout their entire healthcare journey \cite{nguyen2023application}, LLMs are leading medical services towards a more personalized and efficient direction. Leveraging these powerful capacities of LLMs, medics can guarantee that nursing methods are more aligned with patients' unique needs, achieving true patient-centeredness. This technology not only provides precise and detailed medical guidance \cite{walker2023reliability}, enabling interventions to accurately match patients' specific conditions and needs, but also demonstrates remarkable value in clinical practice. Effective utilization of LLMs can improve patient treatment effects as well as prompt healthcare professionals to make more informed decisions based on data, thereby comprehensively improving the quality of patient care. With continuous advancements in LLM technology, its potential to enhance patient care through personalized advice and continuous monitoring is increasingly significant, marking a promising path for modern medicine. Essentially, LLMs represent a significant leap in the field of medical care, reshaping the pattern of patient care by promoting precision medicine, enhancing flexibility, and deepening the patient-centered approach \cite{yang2023exploring}.

\subsubsection{Medical Education}

Medical education \cite{safranek2023role} is rooted in a vast knowledge system and profound practical experience. With their exceptional rapid learning capabilities and multimodal interaction methods, LLMs can efficiently impart professional knowledge to medical personnel, greatly facilitating the retrieval of medical information and the in-depth analysis of cutting-edge research findings. This significantly reduces the time and resource required for healthcare workers to acquire new knowledge. This transformation not only accelerates the training process for high-level medical talent but also strongly promotes innovation and development in the medical field. Meanwhile, medical personnel can use virtual hospital platforms to receive virtual patients, operate mock medical devices, and plan surgical procedures, allowing for thorough rehearsals and preparations before actual operations, further enhancing the quality and efficiency of medical services. Specific examples of how LLMs impact medical education can be considered as follows. Medical educators can use LLMs to create various analogue patient scenes. These scenes fully reproduce the real situation and are very diverse to make sure that students can learn in a real medical environment and accumulate experience through conversations with a large number of patients. For example, medical students can interact with simulated patients with rare diseases, ask questions, and receive answers that mimic real patient responses. Similarly, medical researchers can employ LLMs to swiftly peruse and analyze extensive medical literature, pinpoint pertinent studies, and condense their key findings. This approach dramatically cuts down the time required for conducting literature reviews, thereby enabling researchers to concentrate more on their core research endeavors \cite{karabacak2023advent}.

\subsubsection{Drug Discovery and Development}

LLMs demonstrate potential applications in drug discovery, primarily in analyzing complex molecular structures, screening potential therapeutic compounds, and evaluating the safety and efficacy of drug candidates, thereby contributing to certain aspects of the drug development process \cite{datta2022bert}. Notably, in the cutting-edge field of de novo drug design, chemical language models have achieved remarkable breakthroughs \cite{grisoni2023chemical}. A pioneering study explores strategies for initiating target molecule generation models using pre-trained biochemical language models, comparing direct (one-step) versus phased (two-step) warm-start methods and evaluating the effects of beam search and sampling techniques in compound generation. The results show that models using warm starts exhibit significant advantages over baseline models, with the one-step strategy demonstrating superior versatility and improved performance in docking evaluations and across multiple benchmark indicators. Additionally, beam search technology has been proven to be more efficient and accurate than sampling methods in evaluating the quality of generated compounds \cite{uludougan2022exploiting}. In 2021, the research achieved remarkable results through carefully designed identification experiments targeting protease inhibitors and active site probes using LLMs \cite{hu2021novel}. Meanwhile, experts such as Cloesmeijer ME from the Amsterdam University Medical Centers further explore the broad application prospects of LLM technology in advancing the field of quantitative pharmacology \cite{cloesmeijer2024chatgpt}. Moreover, reinforcement learning algorithms allow models to learn through trial and error in simulated or real environments, maximizing a reward function to optimize treatment regimens, adjust drug dosages, etc., thereby improving treatment outcomes and reducing side effects.

\subsubsection{Medical Information Extraction and Real-time Knowledge Updating}

The application of LLMs in the medical field is also prominent in medical information extraction and knowledge graph construction. Through deep learning and NLP techniques, LLMs can automatically extract key medical entities, attributes, and relationships from vast amounts of unstructured data such as medical literature, medical records, and research reports, providing powerful support for constructing comprehensive and accurate medical knowledge graphs \cite{goel2023llms}. This process not only improves the efficiency of medical information processing but also ensures the timeliness and accuracy of knowledge graphs, providing a solid data foundation for clinical decision support, disease prediction, and treatment optimization. For instance, LLMs can optimize cancer care pathways by dynamically adjusting treatment recommendations based on newly published research and updated therapeutic guidelines \cite{yuan2023advanced}.

\subsubsection{Medical Imaging}

Medical LLMs have demonstrated significant potential in the medical image segmentation field by using deep learning algorithms. By training deep learning models \cite{bhatt2021state}, precise segmentation of tissues, organs, or lesions in medical images can be achieved. These models provide substantial support to radiologists by enabling the early recognition of abnormalities in medical images while providing more precise and detailed diagnostic insights, thereby boosting the accuracy and efficiency of medical imaging analysis \cite{zhao2024chatcad+}. In medical image reconstruction, deep learning algorithms can also help improve the quality of reconstructed images. For example, in CT scans, deep learning algorithms can optimize reconstruction algorithms to reduce noise and artifacts, enhancing image clarity and contrast. In medical imaging, LLMs also hold a pivotal position. For instance, the fusion of LLMs and radionics is applied to predict the treatment outcome of rectal cancer in magnetic resonance imaging (MRI) technology \cite{horvat2022combined}. Some studies have revealed the probability of LLMs to provide precise knowledge on the prevention and screening of breast cancer \cite{haver2023appropriateness}. In the field of radiology, models tailored for specific tasks have been developed, such as accurately extracting pancreatic cystic lesion measurement data from CT and MRI reports \cite{yamashita2021automated}, determining the site of metastatic lesions \cite{do2021patterns}, and rapidly classifying tumor response categories (TRCs) from free-text oncology reports \cite{fink2022deep}.

\subsubsection{Patient Care}

In practical applications, the primary function of LLMs is to answer questions, enabling effective communication with patients \cite{clusmann2023future}. Users can ask LLM questions related to diseases, therapies, lifestyles, and other topics. These models can provide advice that is easy to understand and implement, helping patients better manage their health, enhance their adherence to medical prescriptions, and reduce misunderstandings between doctors and patients, thereby providing better care for patients. EHR plays an important role in nursing work. EHRs serve as tools for documenting comprehensive patient health data, encompassing treatment histories, diagnostic results, therapeutic plans, test findings, and medication specifics. Nurses utilize EHRs to access patient records, document nursing practices, and observations, and collaborate with other healthcare professionals by sharing this data \cite{alzu2021electronic}. LLMs contribute significantly to enhancing the precision and operational efficiency of clinical documentation within EHR systems. These models are capable of autonomously examining and evaluating clinical documents, pinpointing omitted or insufficient data, and furnishing suitable recommendations for enhancement or elucidation. For instance, LLMs can identify discrepancies between diagnostic conclusions and treatment strategies, thereby ensuring that the documented information aligns with the real clinical context.

\subsubsection{Analysis of Medical Literature}

LLMs exhibit significant capability in thoroughly analyzing and succinctly condensing extensive volumes of medical research documents. They can extract key information from literature, such as research methods, experimental results, and inferences. This information can be organized into a structured format, facilitating rapid browsing and understanding by doctors \cite{rathi2024msr28}.    Medical LLMs can automatically classify literature into corresponding medical fields or subfields based on its content, such as clinical medicine, basic medicine, preventive medicine, etc. By summarizing the viewpoints and findings of different literature, medical LLMs can help doctors form a comprehensive understanding of a particular medical issue \cite{han2022analysis}. LLMs can also analyze indicators such as the citation frequency of literature and the impact factor of the publishing journal to initially assess the quality of the literature. This assists doctors in screening out high-quality literature, ensuring that analyses and decisions based on this literature are reliable. In the rapidly evolving field of medicine, keeping up with the latest developments is crucial, and LLMs can help doctors remain at the forefront of innovation in the medical industry and provide evidence-based medical services \cite{tang2023evaluating}. In this regard, healthcare professionals can leverage LLMs to promptly retrieve the latest research outcomes and medical advancements, thereby bypassing the traditional tedious processes associated with online information gathering \cite{dave2023chatgpt}. 

\section{Evaluations of Medical LLMs} \label{evaluation}

Given the highly specialized and sensitive nature of medical data, inaccurate or potentially harmful outputs from medical LLMs could lead to significant patient risks, so rigorous performance evaluation is essential \cite{rydzewski2024comparative}. Our comprehensive analysis categorizes current evaluation approaches for medical LLMs into two principal dimensions: machine evaluation and human-centered evaluation. The former employs quantitative measures to objectively analyze model capabilities, while the latter incorporates human judgment perspectives. Additionally, we classify the studied medical LLMs into three distinct categories based on their training methodologies, as shown in Table \ref{four}.

\begin{table*}[!htp]
    \centering
    \footnotesize
    \caption{Evaluation details of medical LLMs, STQA for single-turn QA, MTQA for multiple-turn QA, MCQA for multiple-choice QA} \label{four}
    \begin{tabular}{|c|c|c|m{2cm}<{\centering}|m{2.5cm}<{\centering}|c|m{3cm}<{\centering}|}
        \hline
         \multirow{2}*{\textbf{Style}} & \multirow{2}*{\textbf{Models}} & \multirow{2}*{\textbf{Year}} & \multirow{2}*{\textbf{Type}} & \multicolumn{1}{c|}{\textbf{Machine evaluation}} & \multicolumn{2}{c|}{\textbf{Human-centric evaluation}}\\
        \cline{5-7}
        & & & & Metrics &  Evaluator & Dimensions\\
        \hline
        \multirow{10}{*}{Pre-training} & GatorTron \cite{yang2022large}& 2022 & STQA & Accuracy,  F1-Score, Precision, Recall & Human & Professional, Safe, Helpful \\
        \cline{2-7}
         & BioGPT \cite{luo2022biogpt}& 2022 & STQA & Accuracy, F1-Score, Precision, Recall & &  \\
         \cline{2-7}
         & ClinicalBERT \cite{alsentzer2019publicly}& 2019 & STQA, MCQA & Accuracy, F1-Score & & \\
         \cline{2-7}
         & BioBERT \cite{lee2020biobert}& 2020 & STQA & Accuracy, F1-Score & Human & Professional, Safe, Helpful  \\
         \cline{2-7}
         & PubMedBERT \cite{gu2021domain}& 2021 & STQA, MCQA & Accuracy, F1-Score & & \\
         \cline{2-7}
         
         & MedFILIP \cite{liang2025medfilip}& 2025 & Non-conversational QA format & ACC, F1-Score & &  \\
        \hline
        \multirow{16}{*}{Fine-tuning} & ChatDoctor \cite{li2023chatdoctor}& 2023 &  STQA & BERTScore & Human & Doctor-like, Patient-friendly, Professional, Safe, Fluent, Helpful\\
        \cline{2-7}

        & HuatuoGPT \cite{zhang2023huatuogpt} & 2023 &  STQA,MTQA & BLEU, ROUGE, GLEU, Distinct  &  Human, LLM & Professional,Proactive, Doctor-like, Patient-friendly \\
        \cline{2-7}
        & BianQue \cite{chen2023bianque}& 2023 & MTQA &  BLEU, ROUGE, PQA & Human & Doctor-like, Patient-friendly, Professional, Proactive  \\
        \cline{2-7}
        &  Zhongjing \cite{yang2024zhongjing}& 2024 &  STQA,MTQA & & Human, LLM & Professional, Safe, Fluent   \\
        \cline{2-7}
        &  Alpacare \cite{zhang2023alpacare}& 2023 & STQA, MTQA & Accuracy, ROUGE & Human, LLM & Professional, Helpful  \\
        \cline{2-7}
        & PMC-LLaMA \cite{wu2024pmc} & 2023 &  MCQA &  Accuracy & & \\
        \cline{2-7}
        & BenTsao \cite{wang2023huatuo} & 2023 & STQA & & Human & Safe, Fluent, Helpful \\
        \cline{2-7}
         &  Qilin-Med \cite{ye2023qilin}& 2023 &  STQA, MTQA & F1-Score, BLEU,  ROUGE & Human & \\
         \cline{2-7}
         &  ClinicalGPT \cite{wang2023clinicalgpt}& 2023 &  STQA, MCQA &  BLEU, ROUGE, GLEU, Accuracy &  LLM & Professional, Safe, Helpful \\
         \cline{2-7}
         & Me-LLaMA \cite{xie2024me} & 2024 &  MCQA, NLP &  Accuracy, F1-score,  ROUGE, BERTScore & & \\
        \hline
        \multirow{10}{*}{Prompting} & Med-PaLM \cite{singhal2025toward}& 2023 & STQA, MCQA & Accuracy & Human & Professional, Safe, Helpful \\
        \cline{2-7}
        & Codex \cite{lievin2024can}& 2024 & MCQA & Accuracy & Human, LLM & Doctor-like, Patient-friendly, Professional, Proactive  \\
        \cline{2-7}
        & MedPrompt \cite{nori2023can}& 2023 & MCQA & Accuracy & Human, LLM & Doctor-like, Professional, Proactive  \\
        \cline{2-7}
        & Chat-Orthopedist \cite{shi2023retrieval}& 2023 & STQA, MCQA & & Human & Doctor-like, Professional, Patient-friendly  \\
        \cline{2-7}
        & DeID-GPT \cite{liu2023deid}& 2023 & STQA & Accuracy & &  \\
        \cline{2-7}
        & AutoMedPrompt \cite{wu2025automedprompt}& 2025 & STQA, MCQA & Accuracy &  & \\
        \hline
    \end{tabular}
 \end{table*}
 
\subsection{Machine Evaluation}

Machine evaluation leverages standardized benchmarks in natural language processing (NLP) to quantitatively analyze medical LLMs' task (handling capabilities). These well-established evaluation frameworks provide clearly defined, uniform metrics for objectively measuring model performance. Most general-purpose LLMs acquire basic linguistic comprehension skills through exposure to language understanding tasks during training. A critical focus of this evaluation paradigm is determining whether medical LLMs retain their foundational language processing competencies after domain-specific adaptation to healthcare data. Illustrative examples include the Me-LLaMA \cite{xie2024me} research initiatives, which utilize NLP benchmarks to test their medical LLM implementations. The methodology typically consists of two phases: specialized supervised fine-tuning using benchmark training data, followed by comprehensive testing on evaluation datasets. Numerous studies employ question-answering tasks (single-turn, multiple-turn, and multiple-choice) as evaluation benchmarks for natural language generation to assess the content quality of responses and the performance of medical LLMs in handling medical QA tasks. The single-turn QA framework involves standalone question-answer pairs without sequential dialogue, enabling direct measurement of a model's proficiency and knowledge breadth on targeted medical queries. This approach accommodates diverse medical topics ranging from basic definitions to intricate diagnostic challenges. Multiple-turn QA benchmarks simulate extended conversational exchanges (e.g., doctor-patient interactions) to evaluate medical LLMs' diagnostic competence, contextual comprehension, and interactive dialogue capabilities. Meanwhile, multiple-choice QA tasks present questions with predefined answer options, allowing precise quantification of model accuracy. Beyond verifying selected answers, some studies further examine the reasoning process behind responses. These varied multiple-choice questions test the model's expertise across medical domains and its capacity for logical deduction. Common evaluation metrics for natural language generation include Accuracy, F1-Score \cite{powers2020evaluation}, BLEU \cite{papineni2002bleu}, ROUGE \cite{lin2004rouge}, and GLEU \cite{wu2016google}. In addition to conventional measures, specialized metrics have been developed for healthcare dialogue contexts. For instance, BianQue \cite{chen2023bianque} introduced Proactive Questioning Ability (PAQ), a novel assessment criterion designed to gauge an LLM's initiative in querying users during interactions.

\subsection{Human-Centered Evaluation}

Due to the diverse and often unpredictable nature of model-generated responses, machine evaluation frequently fails to comprehensively assess medical LLMs’ performance—particularly in critical dimensions such as output safety and practical utility. In response to this challenge, human-centered evaluation has emerged as an increasingly adopted alternative. By incorporating real human judgment and expectations, this approach offers a more realistic and application-oriented assessment of how medical LLMs perform in actual use cases. Current research in the evaluation of medical LLMs frequently employs both human experts and LLMs as assessors. Given that the end users of these systems are human, human-based evaluation remains a directly relevant and effective approach for measuring real-world performance. While human evaluation yields insightful and interpretable assessments of model behavior, it is accompanied by significant limitations—including substantial time investment, high resource demands, and inconsistencies arising from individual biases or varying levels of domain expertise. In response to these challenges, some researchers have turned to advanced LLMs as potential substitutes for human evaluation. These models are trained on vast and diverse datasets, enabling them to develop a knowledge base rivaling that of human experts. Importantly, since such LLMs are designed to mimic human judgment and emulate end-user perspectives, evaluation methodologies based on their output continue to be regarded as part of the human-centric evaluation paradigm. This hybrid strategy aims to balance scalability with the depth of human-like assessment.

\section{Benefits and Drawbacks of Medical LLMs}  
\label{BenefitsandDrawbacks}

\begin{figure}
    \centering
    \includegraphics[scale=0.42]{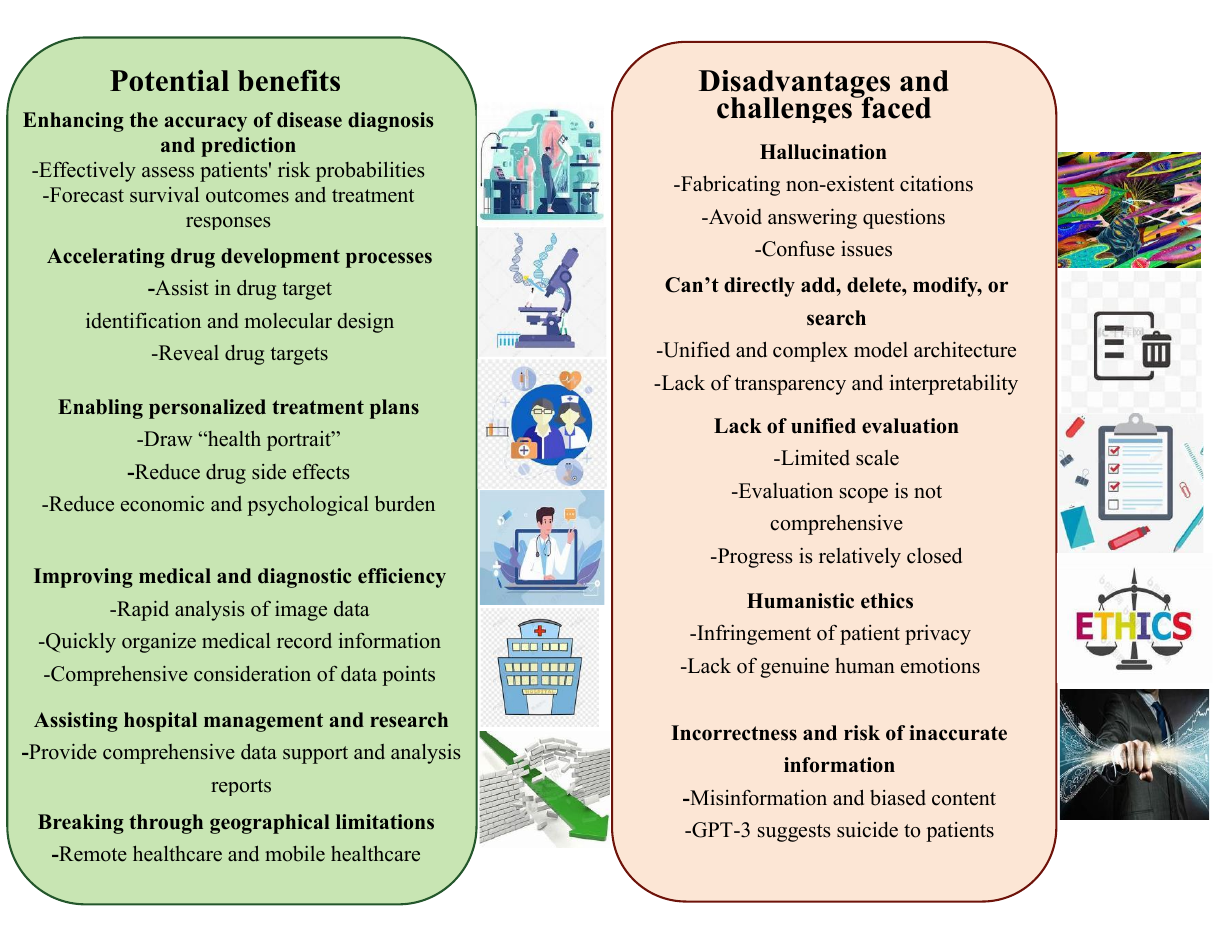}
    \caption{The potential benefits and challenges faced by medical LLMs.}
    \label{fig:pbc}
\end{figure}

When analyzing the dual effects of LLMs in the medical field, we must not only highlight their significant advantages in revolutionizing healthcare models but also address the limitations and challenges they face in integrating into complex medical ecosystems. By comprehensively examining the multiple facets of LLMs, we can more clearly understand their far-reaching impacts on healthcare, including both opportunities for positive change and pressing issues that need to be addressed. The following is a summary analysis of the benefits and challenges of LLMs in medicine, as illustrated in Figure \ref{fig:pbc}.

\subsection{Benefits of Applying Medical LLMs}

\textbf{Enhancing the accuracy of disease diagnosis and prediction.} LLMs play a significant role in early disease detection. By integrating and analyzing clinical diagnostic data \cite{yang2022large} and medical imaging data, this technology can effectively assess patients' risk probabilities for cardiovascular diseases or diabetic complications. Its multidimensional data analysis capability supports the establishment of personalized risk assessment systems, providing a decision-making basis for clinical early intervention. LLMs can also predict and evaluate the efficacy of personalized treatment regimens. Leveraging patients' electronic health records and genomic data, this technology can forecast survival outcomes and treatment responses to specific cancer therapies \cite{iannantuono2023applications}. Through integrated analysis of multi-source data, it enables accurate therapeutic efficacy prediction, providing evidence-based support for clinical decision-making in personalized medicine. This technological advantage contributes to optimizing precision treatment strategies for oncology and other diseases.

\textbf{Accelerating drug development processes.} Drug development has always been a time-consuming and costly endeavor \cite{sinha2018drug}. First, the development cycle is long, as modern biopharmaceuticals require extensive research and development, consuming significant human, material, and financial resources. Second, production is highly complex, involving intricate cell culture and fermentation processes that demand highly specialized technicians and sophisticated equipment. This complexity also makes it difficult to significantly scale up production, limiting the accessibility of drugs. Thirdly, safety control is challenging. Due to the complexity of the components and preparation processes of biopharmaceuticals, controlling safety is more difficult, potentially causing serious side effects and safety issues. The introduction of LLMs has significantly accelerated this process. By simulating the complex interactions between drugs and biomolecules, LLMs can assist in drug target identification and molecular design, providing a scientific basis for drug screening and optimization. This not only shortens the long journey from the lab to the clinic but also significantly reduces trial-and-error costs in the development process \cite{pal2023chatgpt}. More importantly, LLMs can reveal drug targets that are difficult to capture using traditional methods, opening up new directions for new drug development \cite{chakraborty2023artificial}.

\textbf{Enabling personalized treatment plans.} With the advent of the era of precision medicine, personalized treatment plans have become crucial for improving treatment outcomes and reducing patient burdens. LLMs can generate a unique "health portrait" for each patient through a comprehensive analysis of multi-omics data such as the patient's genes, proteome, and metabolome. This "health portrait" includes information on demographic characteristics, physical health, mental health, social health, and other aspects. By using personalized graphics to comprehensively depict and intuitively display the individual's health status, doctors can customize more precise and effective treatment plans, ensuring the pertinence and effectiveness of treatment measures. Personalized treatment not only improves treatment success rates but also reduces unnecessary drug side effects, thereby alleviating patients' economic and psychological burdens and enhancing overall quality of life and satisfaction \cite{stade2024large}. For example, medical professionals use LLMs to develop prevention and management plans for patients with cardiovascular diseases by linking genomic profiles with personal lifestyles \cite{arslan2023exploring}.

\textbf{Improving medical and diagnostic efficiency.} Facing the increasing demand for healthcare, improving medical and diagnostic efficiency is particularly important. LLMs, with their powerful automation capabilities, can undertake some cumbersome medical tasks, such as rapid analysis of imaging data and intelligent organization of medical record information. This not only reduces the workload of healthcare professionals but also makes the medical process smoother and more efficient. For instance, LLMs can predict the likelihood of heart event onset or the risk of diabetes complications for patients through in-depth analysis of vast clinical records and multiple imaging techniques \cite{quer2024potential}. They comprehensively consider various data points to achieve detailed stratification of patient risks and accordingly tailor early intervention plans and rapid diagnoses for each patient. LLMs' exceptional performance in medical image recognition has greatly improved diagnostic efficiency. They can quickly and accurately identify lesion areas and abnormal structures, providing reliable diagnostic evidence and detailed solutions for doctors, minimizing the potential for human error \cite{benary2023leveraging}.

\textbf{Assisting hospital management, teaching, and scientific research.} At the hospital management level, LLMs also play an indispensable role. They provide comprehensive data support and analysis reports for hospital administrators, aiding in the optimal allocation and efficient utilization of medical resources. With the support of LLMs, hospitals can make more precise management decisions, optimize service processes, and improve patient satisfaction. Furthermore, LLMs offer forceful support for medical education and scientific research. They provide abundant data resources and analytical tools to assist researchers in carrying out innovative research work. At the same time, they can also provide vivid cases and practical teaching methods for medical education, nurturing more medical talents with big data thinking and innovative capabilities \cite{zeng2020meddialog}.

\textbf{Breaking geographical limitations and sharing high-quality medical resources.} Geographical limitations have long been one of the significant factors affecting the balanced distribution of medical resources. The introduction of LLMs offers the potential to overcome this limitation. Through the application of new service models such as telemedicine and mobile healthcare, LLMs enable the sharing and dissemination of high-quality medical resources across geographical boundaries. Patients in remote areas can access medical services of the same quality as those in urban areas without traveling long distances. Meanwhile, high-quality medical resources in cities can also be extended to wider regions through the Internet and other channels. This balanced allocation of resources not only improves the accessibility and fairness of medical services but also promotes the overall progress and development of the medical industry.

\subsection{Drawbacks and challenges faced}

\textbf{Hallucinations.} Hallucinations in LLMs refer to the inclusion of inaccurate or factually inconsistent information in their generated outputs. This phenomenon can be subdivided into intrinsic hallucinations and extrinsic hallucinations \cite{rawte2023survey}. Intrinsic hallucinations occur when the generated output contradicts known factual information at the logical level, such as errors in mathematical formula calculations. Extrinsic hallucinations occur when LLM-generated outputs lack verifiable grounding, typically manifesting as fabricated citations or evasive responses. In medical applications, despite their linguistic fluency, such non-factual hallucinations risk propagating misinformation, potentially leading to clinical errors (e.g., misdiagnosis, inappropriate therapies) or misleading patient guidance. For example, when faced with novel queries outside their training distribution, LLMs may conflate concepts and produce medically invalid conclusions.

\textbf{Inability to directly perform CRUD operations.} The core characteristic of LLMs lies in their unified and complex model architecture, which encapsulates both memorized information and learned patterns. This integrated design means that, unlike traditional databases with direct create, read, update, and delete (CRUD) functionalities, one cannot intuitively inspect or modify specific memories or patterns within the model  \cite{amer2023large}. Instead, one can only infer its internal state through continuous question-and-answer interactions, performance evaluations, and indirect reasoning. This limitation results in a lack of transparency and interpretability in the decision-making process of LLMs, raising concerns about safety risks. When updates or adjustments are needed, the inability to directly intervene in the model’s internal mechanisms means that incorporating new knowledge, forgetting outdated information, or fine-tuning behavior can only be achieved by retraining the entire model. This process is not only time-consuming and labor-intensive but may also render the model unable to serve practical applications during the adjustment period, significantly reducing the model's update efficiency and response speed.

\textbf{Lack of unified evaluation} \cite{lin2023llm}.  Currently, although various methods have been adopted to evaluate LLMs, including benchmark testing using public datasets, automated evaluation relying on advanced models such as ChatGPT 4.0, and manual evaluation by professional doctors, these evaluation methods still face numerous challenges. The main issues are their limited scale, incomplete evaluation scope, relatively closed processes, and difficulty in replication in other environments. With the proliferation of generative LLMs, the lack of a universally accepted and standardized evaluation system makes it particularly difficult to compare the performance of different models. This lack of uniform standards not only hinders our objective and comprehensive judgment of model quality but also leads to the isolation of research results, making it difficult to effectively validate and reproduce findings across different studies. These issues ultimately undermine the public's credibility and trust in generative LLMs.

\textbf{Humanistic ethics} \cite{sison2024chatgpt}.  As a rapidly developing emerging field, LLMs are currently facing issues of inadequate industry standards, laws, and regulations. Their construction and training highly depend on vast medical datasets, which often contain highly sensitive personal privacy information such as medical records and genetic information. Once these data are leaked or misused, they may constitute a serious violation of patients' privacy rights. Therefore, protecting privacy is a critical issue when applying LLMs in the medical field \cite{zhui2024ethical}. The handling and management of patients' sensitive information inevitably heighten the risks of data privacy breaches, particularly in cases of unauthorized access or data leaks. When interacting with patients and medical professionals, LLMs may accumulate and store personal health data, including highly sensitive content such as medical records, test results, and diagnostic conclusions. Moreover, although the data collected by LLMs are de-identified, there remains a potential threat of re-identification, as individuals' identities may be restored when these data are cross-referenced with other accessible information sources \cite{el2011systematic}. Thus, maintaining transparency in how patient data is utilized is essential, as patients must be informed about the intended use of their information by LLMs and provided with the opportunity to either consent or decline such utilization. Furthermore, the NLP technologies and machine learning algorithms upon which LLMs depend may introduce risks of privacy violations \cite{ford2020should}. They may inadvertently leak sensitive information or provide inaccurate answers, thereby threatening patients' privacy rights and health status. Despite significant advancements in NLP technologies, LLMs are essentially machine learning models lacking real human emotions and empathy \cite{parviainen2022chatbot}. This results in their responses often appearing cold and mechanical, lacking the warmth characteristic of humans. In a medical environment, patients often desire to establish a warm and compassionate communication bridge with medical personnel, especially when facing health issues or disease challenges \cite{kerasidou2021need}. Therefore, patients are likely to favor LLMs that deliver empathetic and personalized interactions to cater to their emotional needs.

\textbf{Incorrectness and risk of inaccurate information.} LLMs have acquired broad capabilities to perform tasks traditionally requiring human intelligence through training on massive web-based datasets \cite{harrer2023attention}. Nevertheless, such training methods risk internalizing and propagating unverified claims and prejudiced perspectives, which can result in notable drawbacks such as the production of erroneous or entirely invented content \cite{reddy2023evaluating}. The medical and healthcare domains are particularly sensitive to such risks due to their safety-critical nature. Incorrect recommendations concerning patient symptoms and medications could result in severe injury or death \cite{munn2024truth}, as demonstrated by GPT-3's \cite{brown2020language} inappropriate suggestion of suicide to a patient \cite{atallah2023large}. Consequently, implementing robust safeguards becomes imperative for LLM applications in healthcare, particularly when assisting with tasks such as generating discharge summaries, producing automated medical documentation, and providing clinical recommendations.

\section{Future Research Directions} \label{sec:directions}

\begin{figure}
    \centering
    \includegraphics[scale=0.41]{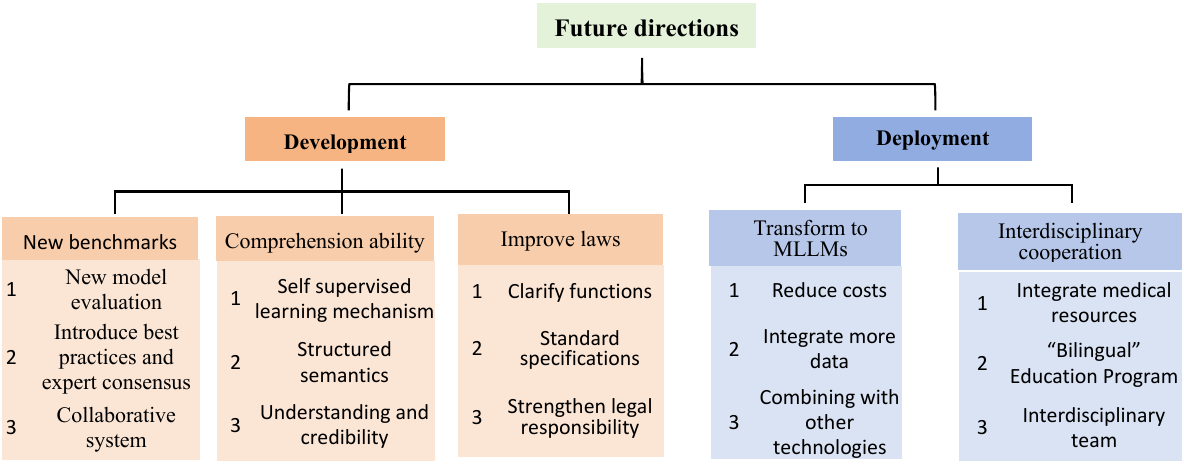}
    \caption{Future research directions.}
    \label{fig:fd}
\end{figure}

It is undeniable that medical LLMs offer extensive opportunities and development directions, such as in the fields of intelligent healthcare \cite{wang2019bio}, research on intelligent or virtual robots \cite{huang2023intelligent,preum2021review}, medical metaverse \cite{chen2022metaverse}, secure healthcare \cite{chen2024metaverse}, blockchain-based healthcare \cite{he2023large}, and collaborations among medical LLMs \cite{roman2018blockchain}. Below are some suggestions for future development, as illustrated in Figure \ref{fig:fd}.

\subsection{Enhancing LLMs' Understanding of Complex Medical Terminology}

Currently, LLMs have achieved remarkable results in natural language processing and understanding, and have also seen rapid development in handling multimodal data \cite{wu2023multimodal}. However, when dealing with specialized data such as gene sequences, protein structures, small molecule structures, chemical equations, pathological images, and analytical reports, these LLMs still have limited understanding capabilities. The medical field is a typical example, filled with a vast amount of complex diagnostic reports, medical images, and medical records, which are characterized by multimodal fusion, complex and varied layouts, a mix of handwritten and printed text, rich numerical expressions, and nested structures. The complexity of medical data is extremely high, which requires us to fully consider its characteristics during both the training and application stages of LLMs. To address this challenge, we first need to delve into the pre-training mechanisms tailored for complex medical data. In particular, we need to design corresponding self-supervised learning mechanisms \cite{mendoza2024adaptive} that incorporate the unique properties of medical data. For instance, in pre-training for protein and small molecule structures, we should fully integrate their chemical and biological characteristics. Additionally, beyond text, we need to design learning mechanisms for document layout, multimodal data, and numerical expressions to guide LLMs in better understanding and processing the semantics of these non-textual complex data. Alignment among different modal data is crucial in this process. Through precise alignment, we can ensure that LLMs accurately understand and integrate information from different modalities, thereby more effectively addressing the complexity of medical data. Future development directions may focus on the following: (i) designing corresponding self-supervised learning mechanisms that incorporate the unique properties of medical data; (ii) incorporating structured semantic knowledge into the training of LLMs; and (iii) improving the comprehensibility and reliability of model outputs.

\subsection{Transitioning LLMs to MLLMs}

Large multimodal models (LMMs), also known as multimodal LLMs (MLLMs) \cite{wu2023multimodal}, have evolved from LLMs to address tasks involving multiple modalities such as vision and text \cite{yin2024survey}. Compared to LLMs, which primarily focus on processing text data, MLLMs exhibit a broader range of capabilities, including the integration of information from multiple modalities, such as patients' medical records, medical images (e.g., MRI, CT, X-rays), and physiological signals. This diverse functionality highlights the significant potential of MLLMs in the medical field. Currently, a series of multimodal LLM frameworks combining vision and language, such as Med-PaLM, LLaVA-Med \cite{li2024llava}, Visual Med-Alpaca \cite{shu2023visual}, Med-Flamingo \cite{moor2023med}, and Qilin-Med-VL \cite{liu2023qilin}, have emerged. These frameworks undergo fine-tuning using paired medical image and text data, significantly enhancing the ability of medical LLMs to understand medical images (e.g., radiological images). The study \cite{huang2023chatgpt} explored the fusion of visual, auditory, and language inputs for automated diagnosis of dental diseases. However, despite the crucial role of time-series data, such as electrocardiograms (ECG) \cite{li2024frozen} and photoplethysmography (PPG) \cite{englhardt2024exploring}, in medical diagnosis and condition monitoring, there are still few medical LLMs capable of handling such data. Although these studies remain preliminary, they indicate that MLLMs may achieve effective cross-domain and cross-modal generalization, extending capabilities beyond traditional NLP tasks. However, due to the high computational costs and limited efficiency of large-scale MLLM training, current models remain significantly smaller in scale compared to conventional LLMs. Future development directions may focus on the following: (i) reducing the costs and improving the efficiency of large-scale training for MLLMs; (ii) incorporating more diverse modal data; (iii) combining with other technologies to further enhance the performance and application scope of MLLMs.

\subsection{Interdisciplinary Collaboration}

Currently, there are significant disciplinary barriers in the medical field, with knowledge, skills, and experiences across different disciplines failing to be effectively integrated, leading to fragmentation in healthcare services. Meanwhile, with the transformation of medical models and the diversification of patient needs, the single-discipline diagnostic and treatment model can no longer meet patients' demands. Therefore, interdisciplinary collaboration in medical LLMs is an inevitable trend in the development of healthcare, integrating knowledge and technologies from computer science, data science, medical expertise, and other disciplines to form a more comprehensive and in-depth medical treatment system. For example, the medical LLM Med-Go \cite{wang2025evaluation}, jointly developed by Shanghai East Hospital and the Institute of Software, Chinese Academy of Sciences, successfully integrated learning and technology from multiple medical disciplines, providing doctors with comprehensive diagnostic tools. This model has passed the National Medical Practitioner Qualification Examination and demonstrated efficient and accurate diagnostic capabilities in multiple real cases. Although interdisciplinary collaboration helps break down barriers between traditional medical disciplines and promotes knowledge exchange and sharing, it still faces several challenges. First, interdisciplinary collaboration involves handling a large amount of medical data that concern patients' privacy and security. To address this issue, strict data protection mechanisms must be established to ensure data security and privacy. Second, differences in terminology and ways of thinking among different disciplines may lead to communication barriers. Effective communication mechanisms should therefore be implemented to facilitate exchanges and cooperation across different disciplines. Furthermore, interdisciplinary collaboration requires the integration of technologies and methods from different disciplines, which may face challenges in technology integration and innovation. To overcome this, it is necessary to strengthen technology research and development, and promote the integration and development of interdisciplinary technologies. Additionally, given the rapid integration of LLMs in healthcare, it becomes increasingly important to cultivate "bilingual" talents who understand both medicine and LLM technology. Future research directions may focus on the following: (i) integrating medical resources from different disciplines to build interdisciplinary knowledge platforms; (ii) launching "bilingual education programs" to provide training in both artificial intelligence and medicine; and (iii) forming interdisciplinary teams to strengthen applied practice.

\subsection{Introducing New Evaluation Benchmarks}

Recent studies have revealed deficiencies in existing benchmarks when assessing the suitability of LLMs for clinical applications \cite{chen2023extensive}. Traditional benchmarks focus on the accuracy of medical question-and-answer sessions, failing to comprehensively evaluate the various skills required of LLMs in clinical practice. Consequently, the use of standardized, human-centered medical exams as an evaluation method for LLMs has been questioned, as passing these exams does not necessarily imply that LLMs possess the complex professional skills needed in real clinical settings \cite{singhal2023large}. To address this issue, a consensus is emerging that more comprehensive benchmarks need to be developed. These benchmarks should encompass abilities such as obtaining information from authoritative medical sources, adapting to changes in the field of medical knowledge, and clearly expressing uncertainty. To enhance the applicability of these benchmarks, future iterations should incorporate real-world clinical simulations to evaluate LLM performance and integrate clinician feedback for continuous refinement, while maintaining robustness. Given the sensitive nature of healthcare applications, the evaluation framework must also rigorously examine ethically critical dimensions—including fairness, bias mitigation, and equitable outcomes—even in the face of measurement complexities \cite{singhal2023large}. Although initiatives such as the AMIE study have utilized evaluations by real physicians and comprehensive criteria grounded in clinical and communication skills to advance benchmarking, there remains a critical need for benchmarks that assess the adaptability, scalability, and robustness of LLMs across diverse and personalized applications \cite{zhou2023survey}. We aim to develop evaluation frameworks that better capture the complexity and variety of practical clinical settings, enabling a more rigorous and context-sensitive assessment of LLMs' applicability in healthcare. Future development directions may focus on the following aspects: (i) integrating simulated environments with real-world data to build a detailed and easily expandable model evaluation framework; (ii) introducing industry best practices and expert consensus to reflect practical application value often overlooked in traditional evaluation systems; (iii) implementing user or expert collaboration evaluation systems to measure LLM performance when incorporating human insight and feedback.

\subsection{Improving Relevant Laws and Regulations}

Currently, LLM technology is undergoing rapid development. However, its application process is accompanied by a series of challenges, including inadequate standardization of application practices, weak awareness of intellectual property protection, and increasingly prominent privacy and ethical issues. In light of this, formulating and improving relevant laws and regulations to strengthen the supervision of LLMs in the medical field is particularly urgent, aiming to guarantee the safety of medical data and the strict protection of patient privacy. Meanwhile, clarifying the responsibility attribution mechanism for the application of technology is crucial to promoting the healthy and efficient development of LLM technology in the medical arena. Notably, some countries and organizations have taken proactive measures. For example, in July 2023, China issued the "Interim Measures for the Administration of Generative AI Services" \cite{zhao2025technological}, providing a legal framework for the standardized development of generative AI services. This suggests that certain governments and institutions are proactively engaging with the challenges associated with the fast advancement of LLMs, seeking to foster the responsible and sustainable progress of medical LLM technology through the establishment and refinement of pertinent policies and regulatory frameworks. Future development directions may focus on the following aspects: (i) clarifying regulatory jurisdictions and enhancing cross-departmental collaboration; (ii) standardizing the application scope and usage standards of medical LLMs; and (iii) strengthening legal responsibilities and accountability mechanisms.

\section{Conclusion}\label{sec:Conclusion}

This article systematically reviews the latest research progress of LLMs in the medical field, thoroughly analyzes their training techniques, medical scenario adaptation strategies, related applications, advantages and limitations of large medical models, and innovatively divides medical big language models into three types based on training methodology. Besides, its evaluation methods are summarized into two categories: machine evaluation and human-centered evaluation. Finally, this study proposes solutions to address the existing challenges in the field of medical LLMs, such as enhancing the ability of medical LLMs to handle complex medical terminology and further developing them into multimodal models. Based on the identified challenges, future research directions are also discussed. By systematically reviewing previous research results, we aim to elucidate the necessity of developing the medical big model, gain a deeper understanding of the current development status of the medical LLMs, and provide directional suggestions for future research.

\section*{Acknowledgment}

This research was supported in part by Guangzhou Basic and Applied Basic Research Foundation (No. 2024A04J9971), National Natural Science Foundation of China (No. 62272196).

\section*{CRediT Authorship Contribution Statement}
Zhiyu Kan: Data curation, Writing original draft.
Wensheng Gan: Review and editing, supervision.
Qilian Qi: Investigation, Review and editing.
Philip S. Yu: Review and editing.

\bibliographystyle{unsrt}

\bibliography{main}

\begin{thebibliography}{100}

\bibitem{minaee2024large}
Shervin Minaee, Tomas Mikolov, Narjes Nikzad, Meysam Chenaghlu, Richard Socher, Xavier Amatriain, and Jianfeng Gao.
\newblock Large language models: A survey.
\newblock {\em arXiv preprint arXiv:2402.06196}, 2024.

\bibitem{radford2018improving}
Alec Radford, Karthik Narasimhan, Tim Salimans, Ilya Sutskever, et~al.
\newblock Improving language understanding by generative pre-training.
\newblock 2018.

\bibitem{gan2023model}
Wensheng Gan, Shicheng Wan, and Phillip~S Yu.
\newblock Model-as-a-service ({MaaS}): A survey.
\newblock In {\em IEEE International Conference on Big Data}, pages 4636--4645. IEEE, 2023.

\bibitem{liu2025screens}
Yihao Liu, Xu~Cao, Tingting Chen, Yankai Jiang, Junjie You, Minghua Wu, Xiaosong Wang, Mengling Feng, Yaochu Jin, and Jintai Chen.
\newblock From screens to scenes: A survey of embodied {AI} in healthcare.
\newblock {\em Information Fusion}, 119:103033, 2025.

\bibitem{swanson1990medical}
Don~R Swanson.
\newblock Medical literature as a potential source of new knowledge.
\newblock {\em Bulletin of the Medical Library Association}, 78(1):29, 1990.

\bibitem{peng2023study}
Cheng Peng, Xi~Yang, Aokun Chen, Kaleb~E Smith, Nima PourNejatian, Anthony~B Costa, Cheryl Martin, Mona~G Flores, Ying Zhang, Tanja Magoc, et~al.
\newblock A study of generative large language model for medical research and healthcare.
\newblock {\em NPJ Digital Medicine}, 6(1):210, 2023.

\bibitem{chow2024unified}
Wei Chow, Juncheng Li, Qifan Yu, Kaihang Pan, Hao Fei, Zhiqi Ge, Shuai Yang, Siliang Tang, Hanwang Zhang, and Qianru Sun.
\newblock Unified generative and discriminative training for multi-modal large language models.
\newblock {\em Advances in Neural Information Processing Systems}, 37:23155--23190, 2024.

\bibitem{popel2018training}
Martin Popel and Ond{\v{r}}ej Bojar.
\newblock Training tips for the {Transformer} model.
\newblock {\em The Prague Bulletin of Mathematical Linguistics}, (110):43--70, 2018.

\bibitem{thirunavukarasu2023large}
Arun~James Thirunavukarasu, Darren Shu~Jeng Ting, Kabilan Elangovan, Laura Gutierrez, Ting~Fang Tan, and Daniel Shu~Wei Ting.
\newblock Large language models in medicine.
\newblock {\em Nature Medicine}, 29(8):1930--1940, 2023.

\bibitem{achiam2023gpt}
Josh Achiam, Steven Adler, Sandhini Agarwal, Lama Ahmad, Ilge Akkaya, Florencia~Leoni Aleman, Diogo Almeida, Janko Altenschmidt, Sam Altman, Shyamal Anadkat, et~al.
\newblock {GPT}-4 technical report.
\newblock {\em arXiv preprint arXiv:2303.08774}, 2023.

\bibitem{santamato2024exploring}
Vito Santamato, Caterina Tricase, Nicola Faccilongo, Massimo Iacoviello, and Agostino Marengo.
\newblock Exploring the impact of artificial intelligence on healthcare management: a combined systematic review and machine-learning approach.
\newblock {\em Applied Sciences}, 14(22):10144, 2024.

\bibitem{karagounis2024leveraging}
Athanasios Karagounis.
\newblock Leveraging large language models for enhancing autonomous vehicle perception.
\newblock {\em arXiv preprint arXiv:2412.20230}, 2024.

\bibitem{pahune2024large}
Saurabh Pahune and Noopur Rewatkar.
\newblock Large language models and generative {AI}'s expanding role in healthcare.
\newblock {\em Preprint}, 2024.

\bibitem{mao2023ad}
Chengsheng Mao, Jie Xu, Luke Rasmussen, Yikuan Li, Prakash Adekkanattu, Jennifer Pacheco, Borna Bonakdarpour, Robert Vassar, Li~Shen, Guoqian Jiang, et~al.
\newblock {AD-BERT}: Using {Pre-trained} language model to predict the progression from mild cognitive impairment to alzheimer's disease.
\newblock {\em Journal of Biomedical Informatics}, 144:104442, 2023.

\bibitem{agbavor2022predicting}
Felix Agbavor and Hualou Liang.
\newblock Predicting dementia from spontaneous speech using large language models.
\newblock {\em PLOS Digital Health}, 1(12):e0000168, 2022.

\bibitem{huang2023chatgpt}
Hanyao Huang, Ou~Zheng, Dongdong Wang, Jiayi Yin, Zijin Wang, Shengxuan Ding, Heng Yin, Chuan Xu, Renjie Yang, Qian Zheng, et~al.
\newblock {ChatGPT} for shaping the future of dentistry: the potential of multi-modal large language model.
\newblock {\em International Journal of Oral Science}, 15(1):29, 2023.

\bibitem{akinci2023large}
Tugba Akinci~D'Antonoli, Arnaldo Stanzione, Christian Bluethgen, Federica Vernuccio, Lorenzo Ugga, Michail~E Klontzas, Renato Cuocolo, Roberto Cannella, and Burak Ko{\c{c}}ak.
\newblock Large language models in radiology: fundamentals, applications, ethical considerations, risks, and future directions.
\newblock {\em Diagnostic and Interventional Radiology}, pages Epub--ahead, 2023.

\bibitem{alberts2023large}
Ian~L Alberts, Lorenzo Mercolli, Thomas Pyka, George Prenosil, Kuangyu Shi, Axel Rominger, and Ali Afshar-Oromieh.
\newblock Large language models {(LLM)} and {ChatGPT}: what will the impact on nuclear medicine be?
\newblock {\em European Journal of Nuclear Medicine and Molecular Imaging}, 50(6):1549--1552, 2023.

\bibitem{singhal2023large}
Karan Singhal, Shekoofeh Azizi, Tao Tu, S~Sara Mahdavi, Jason Wei, Hyung~Won Chung, Nathan Scales, Ajay Tanwani, Heather Cole-Lewis, Stephen Pfohl, et~al.
\newblock Large language models encode clinical knowledge.
\newblock {\em Nature}, 620(7972):172--180, 2023.

\bibitem{su2024based}
Ziqing Su, Guozhang Tang, Rui Huang, Yang Qiao, Zheng Zhang, and Xingliang Dai.
\newblock Based on medicine, the now and future of large language models.
\newblock {\em Cellular and Molecular Bioengineering}, 17(4):263--277, 2024.

\bibitem{ullah2024challenges}
Ehsan Ullah, Anil Parwani, Mirza~Mansoor Baig, and Rajendra Singh.
\newblock Challenges and barriers of using large language models {(LLM)} such as {ChatGPT} for diagnostic medicine with a focus on digital pathology--a recent scoping review.
\newblock {\em Diagnostic Pathology}, 19(1):43, 2024.

\bibitem{bhattacharya2024large}
Manojit Bhattacharya, Soumen Pal, Srijan Chatterjee, Sang-Soo Lee, and Chiranjib Chakraborty.
\newblock Large language model to multimodal large language model: A journey to shape the biological macromolecules to biological sciences and medicine.
\newblock {\em Molecular Therapy-Nucleic Acids}, 35(3), 2024.

\bibitem{waisberg2024concerns}
Ethan Waisberg, Joshua Ong, Mouayad Masalkhi, and Andrew~G Lee.
\newblock Concerns with {OpenAI’s Sora} in medicine.
\newblock {\em Annals of Biomedical Engineering}, pages 1--3, 2024.

\bibitem{shool2025systematic}
Sina Shool, Sara Adimi, Reza Saboori~Amleshi, Ehsan Bitaraf, Reza Golpira, and Mahmood Tara.
\newblock A systematic review of large language model {(LLM)} evaluations in clinical medicine.
\newblock {\em BMC Medical Informatics and Decision Making}, 25(1):117, 2025.

\bibitem{zheng2025large}
Yanxin Zheng, Wensheng Gan, Zefeng Chen, Zhenlian Qi, Qian Liang, and Philip~S Yu.
\newblock Large language models for medicine: a survey.
\newblock {\em International Journal of Machine Learning and Cybernetics}, 16(2):1015--1040, 2025.

\bibitem{majumder2002n}
P~Majumder, M~Mitra, and BB~Chaudhuri.
\newblock {N-gram}: a language independent approach to {IR} and {NLP}.
\newblock In {\em International Conference on Universal Knowledge and Language}, volume~2, 2002.

\bibitem{zhang2010understanding}
Yin Zhang, Rong Jin, and Zhi-Hua Zhou.
\newblock Understanding bag-of-words model: a statistical framework.
\newblock {\em International Journal of Machine Learning and Cybernetics}, 1:43--52, 2010.

\bibitem{bengio2003neural}
Yoshua Bengio, R{\'e}jean Ducharme, Pascal Vincent, and Christian Jauvin.
\newblock A neural probabilistic language model.
\newblock {\em Journal of Machine Learning Research}, 3(Feb):1137--1155, 2003.

\bibitem{jatnika2019word2vec}
Derry Jatnika, Moch~Arif Bijaksana, and Arie~Ardiyanti Suryani.
\newblock {Word2Vec} model analysis for semantic similarities in english words.
\newblock {\em Procedia Computer Science}, 157:160--167, 2019.

\bibitem{devlin2019bert}
Jacob Devlin, Ming-Wei Chang, Kenton Lee, and Kristina Toutanova.
\newblock {BERT}: Pre-training of deep bidirectional transformers for language understanding.
\newblock In {\em The Conference of the North American Chapter of the Association for Computational Linguistics: Human Language Technologies}, pages 4171--4186, 2019.

\bibitem{raffel2020exploring}
Colin Raffel, Noam Shazeer, Adam Roberts, Katherine Lee, Sharan Narang, Michael Matena, Yanqi Zhou, Wei Li, and Peter~J Liu.
\newblock Exploring the limits of transfer learning with a unified text-to-text transformer.
\newblock {\em Journal of Machine Learning Research}, 21(140):1--67, 2020.

\bibitem{gan2025mixture}
Wensheng Gan, Zhenyao Ning, Zhenlian Qi, and Philip~S Yu.
\newblock Mixture of experts ({MOE}): A big data perspective.
\newblock {\em Information Fusion}, pages 1--28, 2025.

\bibitem{bhaskar2023prompted}
Adithya Bhaskar, Alex Fabbri, and Greg Durrett.
\newblock Prompted opinion summarization with {GPT}-3.5.
\newblock In {\em Findings of the Association for Computational Linguistics: ACL}, pages 9282--9300, 2023.

\bibitem{gan2023large}
Wensheng Gan, Zhenlian Qi, Jiayang Wu, and Jerry Chun-Wei Lin.
\newblock Large language models in education: Vision and opportunities.
\newblock In {\em International Conference on Big Data}, pages 4776--4785. IEEE, 2023.

\bibitem{lai2024large}
Jinqi Lai, Wensheng Gan, Jiayang Wu, Zhenlian Qi, and Philip~S Yu.
\newblock Large language models in law: A survey.
\newblock {\em AI Open}, 5:181--196, 2024.

\bibitem{zeng2023large}
Fanlong Zeng, Wensheng Gan, Yongheng Wang, Ning Liu, and Philip~S Yu.
\newblock Large language models for robotics: A survey.
\newblock {\em arXiv preprint arXiv:2311.07226}, 2023.

\bibitem{zeng2023distributed}
Fanlong Zeng, Wensheng Gan, Yongheng Wang, and Philip~S Yu.
\newblock Distributed training of large language models.
\newblock In {\em International Conference on Parallel and Distributed Systems}, pages 840--847. IEEE, 2023.

\bibitem{zeng2025distributed}
Fanlong Zeng, Wensheng Gan, Yongheng Wang, and Philip~S Yu.
\newblock Distributed training of large language models: A survey.
\newblock {\em Natural Language Processing Journal}, page 100174, 2025.

\bibitem{hu2023llm}
Zhiqiang Hu, Lei Wang, Yihuai Lan, Wanyu Xu, Ee-Peng Lim, Lidong Bing, Xing Xu, Soujanya Poria, and Roy Lee.
\newblock {LLM-Adapters}: An adapter family for parameter-efficient fine-tuning of large language models.
\newblock In {\em The Conference on Empirical Methods in Natural Language Processing}, pages 5254--5276, 2023.

\bibitem{liu2023pre}
Pengfei Liu, Weizhe Yuan, Jinlan Fu, Zhengbao Jiang, Hiroaki Hayashi, and Graham Neubig.
\newblock Pre-train, prompt, and predict: A systematic survey of prompting methods in natural language processing.
\newblock {\em ACM Computing Surveys}, 55(9):1--35, 2023.

\bibitem{tom2020brown}
B~Tom et~al.
\newblock Language models are few-shot learners.
\newblock In {\em The 34th International Conference on Neural Information Processing Systems}, pages 1877--1901, 2020.

\bibitem{lee2020biobert}
Jinhyuk Lee, Wonjin Yoon, Sungdong Kim, Donghyeon Kim, Sunkyu Kim, Chan~Ho So, and Jaewoo Kang.
\newblock {BioBERT}: a pre-trained biomedical language representation model for biomedical text mining.
\newblock {\em Bioinformatics}, 36(4):1234--1240, 2020.

\bibitem{gu2021domain}
Yu~Gu, Robert Tinn, Hao Cheng, Michael Lucas, Naoto Usuyama, Xiaodong Liu, Tristan Naumann, Jianfeng Gao, and Hoifung Poon.
\newblock Domain-specific language model pretraining for biomedical natural language processing.
\newblock {\em ACM Transactions on Computing for Healthcare}, 3(1):1--23, 2021.

\bibitem{luo2022biogpt}
Renqian Luo, Liai Sun, Yingce Xia, Tao Qin, Sheng Zhang, Hoifung Poon, and Tie-Yan Liu.
\newblock {BioGPT}: generative pre-trained transformer for biomedical text generation and mining.
\newblock {\em Briefings in Bioinformatics}, 23(6):bbac409, 2022.

\bibitem{labrak2024biomistral}
Yanis Labrak, Adrien Bazoge, Emmanuel Morin, Pierre-antoine Gourraud, Micka{\"e}l Rouvier, and Richard Dufour.
\newblock {BioMistral}: A collection of open-source pretrained large language models for medical domains.
\newblock In {\em 62th Annual Meeting of the Association for Computational Linguistics}, 2024.

\bibitem{alsentzer2019publicly}
Emily Alsentzer, John Murphy, William Boag, Wei-Hung Weng, Di~Jindi, Tristan Naumann, and Matthew McDermott.
\newblock Publicly available clinical.
\newblock In {\em The 2nd Clinical Natural Language Processing Workshop}. ACL, 2019.

\bibitem{singhal2025toward}
Karan Singhal, Tao Tu, Juraj Gottweis, Rory Sayres, Ellery Wulczyn, Mohamed Amin, Le~Hou, Kevin Clark, Stephen~R Pfohl, Heather Cole-Lewis, et~al.
\newblock Toward expert-level medical question answering with large language models.
\newblock {\em Nature Medicine}, pages 1--8, 2025.

\bibitem{toma2023clinical}
Augustin Toma, Patrick~R Lawler, Jimmy Ba, Rahul~G Krishnan, Barry~B Rubin, and Bo~Wang.
\newblock Clinical camel: An open expert-level medical language model with dialogue-based knowledge encoding.
\newblock {\em arXiv preprint arXiv:2305.12031}, 2023.

\bibitem{ouyang2022training}
Long Ouyang, Jeffrey Wu, Xu~Jiang, Diogo Almeida, Carroll Wainwright, Pamela Mishkin, Chong Zhang, Sandhini Agarwal, Katarina Slama, Alex Ray, et~al.
\newblock Training language models to follow instructions with human feedback.
\newblock {\em Advances in Neural Information Processing Systems}, 35:27730--27744, 2022.

\bibitem{jason2022finetuned}
Wei Jason, Bosma Maarten, Zhao Vincent, Guu Kelvin, Adams~Wei Yu, Lester Brian, Nan Du, et~al.
\newblock Finetuned language models are zero-shot learners.
\newblock In {\em International Conference on Learning Representations}, pages 1--46, 2022.

\bibitem{houlsby2019parameter}
Neil Houlsby, Andrei Giurgiu, Stanislaw Jastrzebski, Bruna Morrone, Quentin De~Laroussilhe, Andrea Gesmundo, Mona Attariyan, and Sylvain Gelly.
\newblock Parameter-efficient transfer learning for {NLP}.
\newblock In {\em International Conference on Machine Learning}, pages 2790--2799. PMLR, 2019.

\bibitem{li2023chatdoctor}
Yunxiang Li, Zihan Li, Kai Zhang, Ruilong Dan, Steve Jiang, and You Zhang.
\newblock {ChatDoctor}: A medical chat model fine-tuned on a large language model meta-ai {(LLaMA)} using medical domain knowledge.
\newblock {\em Cureus}, 15(6), 2023.

\bibitem{han2023medalpaca}
Tianyu Han, Lisa~C Adams, Jens-Michalis Papaioannou, Paul Grundmann, Tom Oberhauser, Alexander L{\"o}ser, Daniel Truhn, and Keno~K Bressem.
\newblock {MedAlpaca}--an open-source collection of medical conversational {AI} models and training data.
\newblock {\em arXiv preprint arXiv:2304.08247}, 2023.

\bibitem{ye2023qilin}
Qichen Ye, Junling Liu, Dading Chong, Peilin Zhou, Yining Hua, Fenglin Liu, Meng Cao, Ziming Wang, Xuxin Cheng, Zhu Lei, et~al.
\newblock Qilin-med: Multi-stage knowledge injection advanced medical large language model.
\newblock {\em arXiv preprint arXiv:2310.09089}, 2023.

\bibitem{xiong2023doctorglm}
Honglin Xiong, Sheng Wang, Yitao Zhu, Zihao Zhao, Yuxiao Liu, Linlin Huang, Qian Wang, and Dinggang Shen.
\newblock {DoctorGLM}: Fine-tuning your chinese doctor is not a herculean task.
\newblock {\em arXiv preprint arXiv:2304.01097}, 2023.

\bibitem{du2022glm}
Zhengxiao Du, Yujie Qian, Xiao Liu, Ming Ding, Jiezhong Qiu, Zhilin Yang, and Jie Tang.
\newblock {GLM}: General language model pretraining with autoregressive blank infilling.
\newblock In {\em The 60th Annual Meeting of the Association for Computational Linguistics}, pages 320--335, 2022.

\bibitem{touvron2023llama}
Hugo Touvron, Thibaut Lavril, Gautier Izacard, Xavier Martinet, Marie-Anne Lachaux, Timoth{\'e}e Lacroix, Baptiste Rozi{\`e}re, Naman Goyal, Eric Hambro, Faisal Azhar, et~al.
\newblock {LLaMA}: Open and efficient foundation language models.
\newblock {\em arXiv preprint arXiv:2302.13971}, 2023.

\bibitem{taori2023stanford}
Rohan Taori, Ishaan Gulrajani, Tianyi Zhang, Yann Dubois, Xuechen Li, Carlos Guestrin, Percy Liang, and Tatsunori~B Hashimoto.
\newblock {Stanford Alpaca}: An instruction-following {LLaMA} model, 2023.

\bibitem{yang2024zhongjing}
Songhua Yang, Hanjie Zhao, Senbin Zhu, Guangyu Zhou, Hongfei Xu, Yuxiang Jia, and Hongying Zan.
\newblock Zhongjing: Enhancing the chinese medical capabilities of large language model through expert feedback and real-world multi-turn dialogue.
\newblock In {\em The AAAI Conference on Artificial Intelligence}, volume~38, pages 19368--19376, 2024.

\bibitem{yang2023baichuan}
Aiyuan Yang, Bin Xiao, Bingning Wang, Borong Zhang, Ce~Bian, Chao Yin, Chenxu Lv, Da~Pan, Dian Wang, Dong Yan, et~al.
\newblock Baichuan 2: Open large-scale language models.
\newblock {\em arXiv preprint arXiv:2309.10305}, 2023.

\bibitem{he2025survey}
Kai He, Rui Mao, Qika Lin, Yucheng Ruan, Xiang Lan, Mengling Feng, and Erik Cambria.
\newblock A survey of large language models for healthcare: from data, technology, and applications to accountability and ethics.
\newblock {\em Information Fusion}, 118:102963, 2025.

\bibitem{zhang2024gpt4roi}
Shilong Zhang, Peize Sun, Shoufa Chen, Min Xiao, Wenqi Shao, Wenwei Zhang, Yu~Liu, Kai Chen, and Ping Luo.
\newblock {GPT4RoI}: Instruction tuning large language model on region-of-interest.
\newblock In {\em European Conference on Computer Vision}, pages 52--70. Springer, 2024.

\bibitem{zhang2023alpacare}
Xinlu Zhang, Chenxin Tian, Xianjun Yang, Lichang Chen, Zekun Li, and Linda~Ruth Petzold.
\newblock Alpacare: Instruction-tuned large language models for medical application.
\newblock {\em arXiv preprint arXiv:2310.14558}, 2023.

\bibitem{wang2023huatuo}
Haochun Wang, Chi Liu, Nuwa Xi, Zewen Qiang, Sendong Zhao, Bing Qin, and Ting Liu.
\newblock {HuaTuo}: Tuning llama model with chinese medical knowledge.
\newblock {\em arXiv preprint arXiv:2304.06975}, 2023.

\bibitem{hulora}
Edward~J Hu, Phillip Wallis, Zeyuan Allen-Zhu, Yuanzhi Li, Shean Wang, Lu~Wang, Weizhu Chen, et~al.
\newblock {LoRA}: Low-rank adaptation of large language models.
\newblock In {\em International Conference on Learning Representations}, 2021.

\bibitem{li2021prefix}
Xiang~Lisa Li and Percy Liang.
\newblock Prefix-tuning: Optimizing continuous prompts for generation.
\newblock In {\em The 59th Annual Meeting of the Association for Computational Linguistics and the 11th International Joint Conference on Natural Language Processing}. ACL, 2021.

\bibitem{liu2022p}
Xiao Liu, Kaixuan Ji, Yicheng Fu, Weng Tam, Zhengxiao Du, Zhilin Yang, and Jie Tang.
\newblock P-tuning: Prompt tuning can be comparable to fine-tuning across scales and tasks.
\newblock In {\em The 60th Annual Meeting of the Association for Computational Linguistics}, pages 61--68, 2022.

\bibitem{xu2023baize}
Canwen Xu, Daya Guo, Nan Duan, and Julian McAuley.
\newblock Baize: An open-source chat model with parameter-efficient tuning on self-chat data.
\newblock In {\em The Conference on Empirical Methods in Natural Language Processing}, pages 6268--6278, 2023.

\bibitem{ben2024cpllm}
Ofir Ben~Shoham and Nadav Rappoport.
\newblock {CPLLM}: Clinical prediction with large language models.
\newblock {\em PLOS Digital Health}, 3(12):e0000680, 2024.

\bibitem{chowdhery2023palm}
Aakanksha Chowdhery, Sharan Narang, Jacob Devlin, Maarten Bosma, Gaurav Mishra, Adam Roberts, Paul Barham, Hyung~Won Chung, Charles Sutton, Sebastian Gehrmann, et~al.
\newblock {PaLM}: Scaling language modeling with pathways.
\newblock {\em Journal of Machine Learning Research}, 24(240):1--113, 2023.

\bibitem{wei2022chain}
Jason Wei, Xuezhi Wang, Dale Schuurmans, Maarten Bosma, Fei Xia, Ed~Chi, Quoc~V Le, Denny Zhou, et~al.
\newblock {Chain-of-Thought} prompting elicits reasoning in large language models.
\newblock {\em Advances in Neural Information Processing Systems}, 35:24824--24837, 2022.

\bibitem{lester2021power}
Brian Lester, Rami Al-Rfou, and Noah Constant.
\newblock The power of scale for parameter-efficient prompt tuning.
\newblock In {\em The Conference on Empirical Methods in Natural Language Processing}. ACL, 2021.

\bibitem{lewis2020retrieval}
Patrick Lewis, Ethan Perez, Aleksandra Piktus, Fabio Petroni, Vladimir Karpukhin, Naman Goyal, Heinrich K{\"u}ttler, Mike Lewis, Wen-tau Yih, Tim Rockt{\"a}schel, et~al.
\newblock Retrieval-augmented generation for knowledge-intensive {NLP} tasks.
\newblock {\em Advances in Neural Information Processing Systems}, 33:9459--9474, 2020.

\bibitem{tang2024instance}
Long Tang, Jingtao Zhao, Yingjie Tian, Changhua Yao, and Panos~M Pardalos.
\newblock Instance-wise multi-view visual fusion for zero-shot learning.
\newblock {\em Applied Soft Computing}, 167:112339, 2024.

\bibitem{yuan2024generalized}
Hao-tian Yuan, Ke-kun Huang, Jie-li Duan, Li-qian Lai, Jia-xiang Yu, Chao-wei Huang, and Zhou Yang.
\newblock Generalized few-shot learning for crop hyperspectral image precise classification.
\newblock {\em Computers and Electronics in Agriculture}, 227:109498, 2024.

\bibitem{liu2023deid}
Zhengliang Liu, Yue Huang, Xiaowei Yu, Lu~Zhang, Zihao Wu, Chao Cao, Haixing Dai, Lin Zhao, Yiwei Li, Peng Shu, et~al.
\newblock {DeID-GPT}: Zero-shot medical text de-identification by {GPT}-4.
\newblock {\em arXiv e-prints}, pages arXiv--2303, 2023.

\bibitem{nori2023can}
Harsha Nori, Yin~Tat Lee, Sheng Zhang, Dean Carignan, Richard Edgar, Nicolo Fusi, Nicholas King, Jonathan Larson, Yuanzhi Li, Weishung Liu, et~al.
\newblock Can generalist foundation models outcompete special-purpose tuning? case study in medicine.
\newblock {\em arXiv preprint arXiv:2311.16452}, 2023.

\bibitem{gao2023retrieval}
Yunfan Gao, Yun Xiong, Xinyu Gao, Kangxiang Jia, Jinliu Pan, Yuxi Bi, Yixin Dai, Jiawei Sun, Haofen Wang, and Haofen Wang.
\newblock Retrieval-augmented generation for large language models: A survey.
\newblock {\em arXiv preprint arXiv:2312.10997}, 2(1), 2023.

\bibitem{xiong2024benchmarking}
Guangzhi Xiong, Qiao Jin, Zhiyong Lu, and Aidong Zhang.
\newblock Benchmarking retrieval-augmented generation for medicine.
\newblock In {\em Findings of the Association for Computational Linguistics ACL}, pages 6233--6251, 2024.

\bibitem{li2023angle}
Xianming Li and Jing Li.
\newblock Angle-optimized text embeddings.
\newblock {\em arXiv e-prints}, pages arXiv--2309, 2023.

\bibitem{wang2023voyager}
Guanzhi Wang, Yuqi Xie, Yunfan Jiang, Ajay Mandlekar, Chaowei Xiao, Yuke Zhu, Linxi Fan, and Anima Anandkumar.
\newblock {VOYAGER}: An open-ended embodied agent with large language models.
\newblock {\em arXiv preprint arXiv:2305.16291}, 2023.

\bibitem{xiao2024c}
Shitao Xiao, Zheng Liu, Peitian Zhang, Niklas Muennighoff, Defu Lian, and Jian-Yun Nie.
\newblock {C-Pack}: Packed resources for general chinese embeddings.
\newblock In {\em The 47th International ACM SIGIR Conference on Research and Development in Information Retrieval}, pages 641--649, 2024.

\bibitem{shao2023enhancing}
Zhihong Shao, Yeyun Gong, Yelong Shen, Minlie Huang, Nan Duan, and Weizhu Chen.
\newblock Enhancing retrieval-augmented large language models with iterative retrieval-generation synergy.
\newblock In {\em Findings of the Association for Computational Linguistics: EMNLP 2023}, pages 9248--9274, 2023.

\bibitem{trivedi2023interleaving}
Harsh Trivedi, Niranjan Balasubramanian, Tushar Khot, and Ashish Sabharwal.
\newblock Interleaving retrieval with chain-of-thought reasoning for knowledge-intensive multi-step questions.
\newblock In {\em The 61st Annual Meeting of the Association for Computational Linguistics}, 2023.

\bibitem{asai2024self}
Akari Asai, Zeqiu Wu, Yizhong Wang, Avi Sil, and Hannaneh Hajishirzi.
\newblock {Self-RAG}: Learning to retrieve, generate, and critique through self-reflection.
\newblock In {\em International Conference on Learning Representations}, 2024.

\bibitem{wu2024clinical}
Jiageng Wu, Xiaocong Liu, Minghui Li, Wanxin Li, Zichang Su, Shixu Lin, Lucas Garay, Zhiyun Zhang, Yujie Zhang, Qingcheng Zeng, et~al.
\newblock Clinical text datasets for medical artificial intelligence and large language models—a systematic review.
\newblock {\em NEJM AI}, 1(6):AIra2400012, 2024.

\bibitem{genai2023llama}
Meta GenAI.
\newblock {LLaMA} 2: Open foundation and fine-tuned chat models.
\newblock {\em arXiv preprint arXiv:2307.09288}, 2023.

\bibitem{shore2009art}
Carol Shore.
\newblock The art of healing and the science of art.
\newblock {\em Healing with Art and Soul: Engaging One’s Self Through Art modalities}, pages 2--13, 2009.

\bibitem{ackerknecht2016short}
Erwin~H Ackerknecht.
\newblock {\em A short history of medicine}.
\newblock JHU press, 2016.

\bibitem{castiglioni2019history}
Arturo Castiglioni.
\newblock {\em A history of medicine}.
\newblock Routledge, 2019.

\bibitem{lammert2024expert}
Jacqueline Lammert, Tobias Dreyer, Sonja Mathes, Leonid Kuligin, Kai~J Borm, Ulrich~A Schatz, Marion Kiechle, Alisa~M L{\"o}rsch, Johannes Jung, Sebastian Lange, et~al.
\newblock Expert-guided large language models for clinical decision support in precision oncology.
\newblock {\em JCO Precision Oncology}, 8:e2400478, 2024.

\bibitem{radford2019language}
Alec Radford, Jeffrey Wu, Rewon Child, David Luan, Dario Amodei, Ilya Sutskever, et~al.
\newblock Language models are unsupervised multitask learners.
\newblock {\em OpenAI Blog}, 1(8):9, 2019.

\bibitem{brown2020language}
Tom Brown, Benjamin Mann, Nick Ryder, Melanie Subbiah, Jared~D Kaplan, Prafulla Dhariwal, Arvind Neelakantan, Pranav Shyam, Girish Sastry, Amanda Askell, et~al.
\newblock Language models are few-shot learners.
\newblock {\em Advances in Neural Information Processing Systems}, 33:1877--1901, 2020.

\bibitem{lin2021m6}
Junyang Lin, An~Yang, Jinze Bai, Chang Zhou, Le~Jiang, Xianyan Jia, Ang Wang, Jie Zhang, Yong Li, Wei Lin, et~al.
\newblock {M6-10T}: A sharing-delinking paradigm for efficient multi-trillion parameter pretraining.
\newblock {\em arXiv preprint arXiv:2110.03888}, 2021.

\bibitem{ling2023domain}
Chen Ling, Xujiang Zhao, Jiaying Lu, Chengyuan Deng, Can Zheng, Junxiang Wang, Tanmoy Chowdhury, Yun Li, Hejie Cui, Xuchao Zhang, et~al.
\newblock Domain specialization as the key to make large language models disruptive: A comprehensive survey.
\newblock {\em ACM Computing Surveys}, 2025.

\bibitem{bedi2024testing}
Suhana Bedi, Yutong Liu, Lucy Orr-Ewing, Dev Dash, Sanmi Koyejo, Alison Callahan, Jason~A Fries, Michael Wornow, Akshay Swaminathan, Lisa~Soleymani Lehmann, et~al.
\newblock Testing and evaluation of health care applications of large language models: a systematic review.
\newblock {\em JAMA}, 2024.

\bibitem{sutton2020overview}
Reed~T Sutton, David Pincock, Daniel~C Baumgart, Daniel~C Sadowski, Richard~N Fedorak, and Karen~I Kroeker.
\newblock An overview of clinical decision support systems: benefits, risks, and strategies for success.
\newblock {\em NPJ Digital Medicine}, 3(1):17, 2020.

\bibitem{hovy1987generating}
Eduard Hovy.
\newblock Generating natural language under pragmatic constraints.
\newblock {\em Journal of Pragmatics}, 11(6):689--719, 1987.

\bibitem{aryan2023costly}
Abi Aryan, Aakash~Kumar Nain, Andy McMahon, Lucas~Augusto Meyer, and Harpreet~Singh Sahota.
\newblock The costly dilemma: are large language models the pay-day loans of machine learning, 2023.

\bibitem{xia2024understanding}
Yuchen Xia, Jiho Kim, Yuhan Chen, Haojie Ye, Souvik Kundu, Cong~Callie Hao, and Nishil Talati.
\newblock Understanding the performance and estimating the cost of {LLM} fine-tuning.
\newblock In {\em International Symposium on Workload Characterization}, pages 210--223. IEEE, 2024.

\bibitem{lin2015rwfim}
Jerry Chun-Wei Lin, Wensheng Gan, Philippe Fournier-Viger, and Tzung-Pei Hong.
\newblock {RWFIM}: Recent weighted-frequent itemsets mining.
\newblock {\em Engineering Applications of Artificial Intelligence}, 45:18--32, 2015.

\bibitem{gan2023anomaly}
Wensheng Gan, Lili Chen, Shicheng Wan, Jiahui Chen, and Chien-Ming Chen.
\newblock Anomaly rule detection in sequence data.
\newblock {\em IEEE Transactions on Knowledge and Data Engineering}, 35(12):12095--12108, 2023.

\bibitem{chen2023language}
Zekai Chen, Mariann~Micsinai Balan, and Kevin Brown.
\newblock Language models are few-shot learners for prognostic prediction.
\newblock {\em arXiv preprint arXiv:2302.12692}, 2023.

\bibitem{xue2023potential}
Vivian~Weiwen Xue, Pinggui Lei, and William~C Cho.
\newblock The potential impact of {ChatGPT} in clinical and translational medicine.
\newblock {\em Clinical and Translational Medicine}, 13(3), 2023.

\bibitem{chen2023boosting}
Zekai Chen, Mariann~Micsinai Balan, and Kevin Brown.
\newblock Boosting transformers and language models for clinical prediction in immunotherapy.
\newblock In {\em The 61st Annual Meeting of the Association for Computational Linguistics}, pages 332--340, 2023.

\bibitem{adoma2020comparative}
Acheampong~Francisca Adoma, Nunoo-Mensah Henry, and Wenyu Chen.
\newblock Comparative analyses of {BERT}, {RoBERTa}, {DistilBERT}, and xlnet for text-based emotion recognition.
\newblock In {\em 17th International Computer Conference on Wavelet Active Media Technology and Information Processing}, pages 117--121. IEEE, 2020.

\bibitem{malla2021covid}
SreeJagadeesh Malla and PJA Alphonse.
\newblock {COVID-19} outbreak: An ensemble pre-trained deep learning model for detecting informative tweets.
\newblock {\em Applied Soft Computing}, 107:107495, 2021.

\bibitem{li2023text}
Hongyang Li, Richard~C Gerkin, Alyssa Bakke, Raquel Norel, Guillermo Cecchi, Christophe Laudamiel, Masha~Y Niv, Kathrin Ohla, John~E Hayes, Valentina Parma, et~al.
\newblock Text-based predictions of {COVID-19} diagnosis from self-reported chemosensory descriptions.
\newblock {\em Communications Medicine}, 3(1):104, 2023.

\bibitem{bill2023fine}
Desir{\'e}e Bill and Theodor Eriksson.
\newblock Fine-tuning a {LLM} using reinforcement learning from human feedback for a therapy chatbot application, 2023.

\bibitem{bilal2024enhancing}
Maham Bilal, Yumna Jamil, Dua Rana, and Hussain~Haider Shah.
\newblock Enhancing awareness and self-diagnosis of obstructive sleep apnea using ai-powered chatbots: the role of {ChatGPT} in revolutionizing healthcare.
\newblock {\em Annals of Biomedical Engineering}, 52(2):136--138, 2024.

\bibitem{javaid2023chatgpt}
Mohd Javaid, Abid Haleem, and Ravi~Pratap Singh.
\newblock {ChatGPT} for healthcare services: An emerging stage for an innovative perspective.
\newblock {\em BenchCouncil Transactions on Benchmarks, Standards and Evaluations}, 3(1):100105, 2023.

\bibitem{eapen2023personalization}
Joel Eapen and VS~Adhithyan.
\newblock Personalization and customization of {LLM} responses.
\newblock {\em International Journal of Research Publication and Reviews}, 4(12):2617--2627, 2023.

\bibitem{nguyen2023application}
Josh Nguyen and Christopher~A Pepping.
\newblock The application of {ChatGPT} in healthcare progress notes: a commentary from a clinical and research perspective.
\newblock {\em Clinical and Translational Medicine}, 13(7), 2023.

\bibitem{walker2023reliability}
Harriet~Louise Walker, Shahi Ghani, Christoph Kuemmerli, Christian~Andreas Nebiker, Beat~Peter M{\"u}ller, Dimitri~Aristotle Raptis, and Sebastian~Manuel Staubli.
\newblock Reliability of medical information provided by {ChatGPT}: assessment against clinical guidelines and patient information quality instrument.
\newblock {\em Journal of Medical Internet Research}, 25:e47479, 2023.

\bibitem{yang2023exploring}
Hao Yang, Jiaxi Li, Siru Liu, Lei Du, Xiali Liu, Yong Huang, Qingke Shi, and Jialin Liu.
\newblock Exploring the potential of large language models in personalized diabetes treatment strategies.
\newblock {\em MedRxiv}, pages 2023--06, 2023.

\bibitem{safranek2023role}
Conrad~W Safranek, Anne~Elizabeth Sidamon-Eristoff, Aidan Gilson, and David Chartash.
\newblock The role of large language models in medical education: applications and implications, 2023.

\bibitem{karabacak2023advent}
Mert Karabacak, Burak~Berksu Ozkara, Konstantinos Margetis, Max Wintermark, and Sotirios Bisdas.
\newblock The advent of generative language models in medical education.
\newblock {\em JMIR Medical Education}, 9:e48163, 2023.

\bibitem{datta2022bert}
Tanmoy~Tapos Datta, Pintu~Chandra Shill, and Zabir Al~Nazi.
\newblock {BERT-D2}: Drug-drug interaction extraction using {BERT}.
\newblock In {\em International Conference for Advancement in Technology}, pages 1--6. IEEE, 2022.

\bibitem{grisoni2023chemical}
Francesca Grisoni.
\newblock Chemical language models for de novo drug design: Challenges and opportunities.
\newblock {\em Current Opinion in Structural Biology}, 79:102527, 2023.

\bibitem{uludougan2022exploiting}
G{\"o}k{\c{c}}e Uludo{\u{g}}an, Elif Ozkirimli, Kutlu~O Ulgen, Nilg{\"u}n Karal{\i}, and Arzucan {\"O}zg{\"u}r.
\newblock Exploiting pretrained biochemical language models for targeted drug design.
\newblock {\em Bioinformatics}, 38(Supplement\_2):ii155--ii161, 2022.

\bibitem{hu2021novel}
Fan Hu, Lei Wang, Yishen Hu, Dongqi Wang, Weijie Wang, Jianbing Jiang, Nan Li, and Peng Yin.
\newblock A novel framework integrating {AI} model and enzymological experiments promotes identification of {SARS-CoV-2 3CL} protease inhibitors and activity-based probe.
\newblock {\em Briefings in Bioinformatics}, 22(6):bbab301, 2021.

\bibitem{cloesmeijer2024chatgpt}
Michael~E Cloesmeijer, Alexander Janssen, Sjoerd~F Koopman, Marjon~H Cnossen, Ron~AA Math{\^o}t, and SYMPHONY consortium.
\newblock {ChatGPT} in pharmacometrics? potential opportunities and limitations.
\newblock {\em British Journal of Clinical Pharmacology}, 90(1):360--365, 2024.

\bibitem{goel2023llms}
Akshay Goel, Almog Gueta, Omry Gilon, Chang Liu, Sofia Erell, Lan~Huong Nguyen, Xiaohong Hao, Bolous Jaber, Shashir Reddy, Rupesh Kartha, et~al.
\newblock {LLMs} accelerate annotation for medical information extraction.
\newblock In {\em Machine Learning for Health}, pages 82--100. PMLR, 2023.

\bibitem{yuan2023advanced}
J~Yuan, P~Bao, Z~Chen, M~Yuan, J~Zhao, J~Pan, Y~Xie, Y~Cao, Y~Wang, Z~Wang, et~al.
\newblock Advanced prompting as a catalyst: Empowering large language models in the management of gastrointestinal cancers.
\newblock {\em The Innovation}, 521, 2023.

\bibitem{bhatt2021state}
Chandradeep Bhatt, Indrajeet Kumar, V~Vijayakumar, Kamred~Udham Singh, and Abhishek Kumar.
\newblock The state of the art of deep learning models in medical science and their challenges.
\newblock {\em Multimedia Systems}, 27(4):599--613, 2021.

\bibitem{zhao2024chatcad+}
Zihao Zhao, Sheng Wang, Jinchen Gu, Yitao Zhu, Lanzhuju Mei, Zixu Zhuang, Zhiming Cui, Qian Wang, and Dinggang Shen.
\newblock {ChatCAD+}: Toward a universal and reliable interactive cad using llms.
\newblock {\em IEEE Transactions on Medical Imaging}, 43(11):3755--3766, 2024.

\bibitem{horvat2022combined}
Natally Horvat, Harini Veeraraghavan, Caio~SR Nahas, David~DB Bates, Felipe~R Ferreira, Junting Zheng, Marinela Capanu, James~L Fuqua~III, Maria~Clara Fernandes, Ramon~E Sosa, et~al.
\newblock Combined artificial intelligence and radiologist model for predicting rectal cancer treatment response from magnetic resonance imaging: an external validation study.
\newblock {\em Abdominal Radiology}, 47(8):2770--2782, 2022.

\bibitem{haver2023appropriateness}
Hana~L Haver, Emily~B Ambinder, Manisha Bahl, Eniola~T Oluyemi, Jean Jeudy, and Paul~H Yi.
\newblock Appropriateness of breast cancer prevention and screening recommendations provided by {ChatGPT}.
\newblock {\em Radiology}, 307(4):e230424, 2023.

\bibitem{yamashita2021automated}
Rikiya Yamashita, Kristen Bird, Philip Yue-Cheng Cheung, Johannes~Hugo Decker, Marta~Nicole Flory, Daniel Goff, Linda~Nayeli Morimoto, Andy Shon, Andrew~Louis Wentland, Daniel~L Rubin, et~al.
\newblock Automated identification and measurement extraction of pancreatic cystic lesions from free-text radiology reports using natural language processing.
\newblock {\em Radiology: Artificial Intelligence}, 4(2):e210092, 2021.

\bibitem{do2021patterns}
Richard~KG Do, Kaelan Lupton, Pamela~I Causa~Andrieu, Anisha Luthra, Michio Taya, Karen Batch, Huy Nguyen, Prachi Rahurkar, Lior Gazit, Kevin Nicholas, et~al.
\newblock Patterns of metastatic disease in patients with cancer derived from natural language processing of structured {CT} radiology reports over a 10-year period.
\newblock {\em Radiology}, 301(1):115--122, 2021.

\bibitem{fink2022deep}
Matthias~A Fink, Klaus Kades, Arved Bischoff, Martin Moll, Merle Schnell, Maike K{\"u}chler, Gregor K{\"o}hler, Jan Sellner, Claus~Peter Heussel, Hans-Ulrich Kauczor, et~al.
\newblock Deep learning--based assessment of oncologic outcomes from natural language processing of structured radiology reports.
\newblock {\em Radiology: Artificial Intelligence}, 4(5):e220055, 2022.

\bibitem{clusmann2023future}
Jan Clusmann, Fiona~R Kolbinger, Hannah~Sophie Muti, Zunamys~I Carrero, Jan-Niklas Eckardt, Narmin~Ghaffari Laleh, Chiara Maria~Lavinia L{\"o}ffler, Sophie-Caroline Schwarzkopf, Michaela Unger, Gregory~P Veldhuizen, et~al.
\newblock The future landscape of large language models in medicine.
\newblock {\em Communications Medicine}, 3(1):141, 2023.

\bibitem{alzu2021electronic}
Amal~A Alzu'bi, Valerie~JM Watzlaf, and Patty Sheridan.
\newblock Electronic health record {(EHR)} abstraction.
\newblock {\em Perspectives in Health Information Management}, 18(Spring), 2021.

\bibitem{rathi2024msr28}
H~Rathi, A~Malik, DC~Behera, and G~Kamboj.
\newblock {MSR28} use of large language model {(LLM)} for full-text screening in systematic literature reviews: A comparative analysis.
\newblock {\em Value in Health}, 27(6):S264, 2024.

\bibitem{han2022analysis}
Jeong-Won Han, Junhee Park, and Hanna Lee.
\newblock Analysis of the effect of an artificial intelligence chatbot educational program on non-face-to-face classes: a quasi-experimental study.
\newblock {\em BMC Medical Education}, 22(1):830, 2022.

\bibitem{tang2023evaluating}
Liyan Tang, Zhaoyi Sun, Betina Idnay, Jordan~G Nestor, Ali Soroush, Pierre~A Elias, Ziyang Xu, Ying Ding, Greg Durrett, Justin~F Rousseau, et~al.
\newblock Evaluating large language models on medical evidence summarization.
\newblock {\em NPJ Digital Medicine}, 6(1):158, 2023.

\bibitem{dave2023chatgpt}
Tirth Dave, Sai~Anirudh Athaluri, and Satyam Singh.
\newblock {ChatGPT} in medicine: an overview of its applications, advantages, limitations, future prospects, and ethical considerations.
\newblock {\em Frontiers in Artificial Intelligence}, 6:1169595, 2023.

\bibitem{rydzewski2024comparative}
Nicholas~R Rydzewski, Deepak Dinakaran, Shuang~G Zhao, Eytan Ruppin, Baris Turkbey, Deborah~E Citrin, and Krishnan~R Patel.
\newblock Comparative evaluation of {LLMs} in clinical oncology.
\newblock {\em Nejm AI}, 1(5):AIoa2300151, 2024.

\bibitem{yang2022large}
Xi~Yang, Aokun Chen, Nima PourNejatian, Hoo~Chang Shin, Kaleb~E Smith, Christopher Parisien, Colin Compas, Cheryl Martin, Anthony~B Costa, Mona~G Flores, et~al.
\newblock A large language model for electronic health records.
\newblock {\em NPJ Digital Medicine}, 5(1):194, 2022.

\bibitem{liang2025medfilip}
Xinjie Liang, Xiangyu Li, Fanding Li, Jie Jiang, Qing Dong, Wei Wang, Kuanquan Wang, Suyu Dong, Gongning Luo, and Shuo Li.
\newblock {MedFILIP}: Medical fine-grained language-image pre-training.
\newblock {\em IEEE Journal of Biomedical and Health Informatics}, 2025.

\bibitem{zhang2023huatuogpt}
Hongbo Zhang, Junying Chen, Feng Jiang, Fei Yu, Zhihong Chen, Guiming Chen, Jianquan Li, Xiangbo Wu, Zhang Zhiyi, Qingying Xiao, et~al.
\newblock {HuatuoGPT}, towards taming language model to be a doctor.
\newblock In {\em Findings of the Association for Computational Linguistics: EMNLP}, pages 10859--10885, 2023.

\bibitem{chen2023bianque}
Yirong Chen, Zhenyu Wang, Xiaofen Xing, Zhipei Xu, Kai Fang, Junhong Wang, Sihang Li, Jieling Wu, Qi~Liu, Xiangmin Xu, et~al.
\newblock {BianQue}: Balancing the questioning and suggestion ability of health llms with multi-turn health conversations polished by {ChatGPT}.
\newblock {\em arXiv preprint arXiv:2310.15896}, 2023.

\bibitem{wu2024pmc}
Chaoyi Wu, Weixiong Lin, Xiaoman Zhang, Ya~Zhang, Weidi Xie, and Yanfeng Wang.
\newblock {PMC-LLaMA}: toward building open-source language models for medicine.
\newblock {\em Journal of the American Medical Informatics Association}, 31(9):1833--1843, 2024.

\bibitem{wang2023clinicalgpt}
Guangyu Wang, Guoxing Yang, Zongxin Du, Longjun Fan, and Xiaohu Li.
\newblock {ClinicalGPT}: large language models finetuned with diverse medical data and comprehensive evaluation.
\newblock {\em arXiv preprint arXiv:2306.09968}, 2023.

\bibitem{xie2024me}
Qianqian Xie, Qingyu Chen, Aokun Chen, Cheng Peng, Yan Hu, Fongci Lin, Xueqing Peng, Jimin Huang, Jeffrey Zhang, Vipina Keloth, et~al.
\newblock Me-{LLaMA}: Foundation large language models for medical applications.
\newblock {\em Research Square}, pages rs--3, 2024.

\bibitem{lievin2024can}
Valentin Li{\'e}vin, Christoffer~Egeberg Hother, Andreas~Geert Motzfeldt, and Ole Winther.
\newblock Can large language models reason about medical questions?
\newblock {\em Patterns}, 5(3), 2024.

\bibitem{shi2023retrieval}
Wenqi Shi, Yuchen Zhuang, Yuanda Zhu, Henry Iwinski, Michael Wattenbarger, and May~Dongmei Wang.
\newblock Retrieval-augmented large language models for adolescent idiopathic scoliosis patients in shared decision-making.
\newblock In {\em The 14th ACM International Conference on Bioinformatics, Computational Biology, and Health Informatics}, pages 1--10, 2023.

\bibitem{wu2025automedprompt}
Sean Wu, Michael Koo, Fabien Scalzo, and Ira Kurtz.
\newblock {AutoMedPrompt}: A new framework for optimizing {LLM} medical prompts using textual gradients.
\newblock {\em arXiv preprint arXiv:2502.15944}, 2025.

\bibitem{powers2020evaluation}
David~MW Powers.
\newblock Evaluation: from precision, recall and f-measure to {ROC}, informedness, markedness and correlation.
\newblock {\em arXiv preprint arXiv:2010.16061}, 2020.

\bibitem{papineni2002bleu}
Kishore Papineni, Salim Roukos, Todd Ward, and Wei-Jing Zhu.
\newblock {BLEU}: a method for automatic evaluation of machine translation.
\newblock In {\em The 40th Annual Meeting of the Association for Computational Linguistics}, pages 311--318, 2002.

\bibitem{lin2004rouge}
Chin-Yew Lin.
\newblock {ROUGE}: A package for automatic evaluation of summaries.
\newblock In {\em Text Summarization Branches Out}, pages 74--81, 2004.

\bibitem{wu2016google}
Yonghui Wu, Mike Schuster, Zhifeng Chen, Quoc~V Le, Mohammad Norouzi, Wolfgang Macherey, Maxim Krikun, Yuan Cao, Qin Gao, Klaus Macherey, et~al.
\newblock Google's neural machine translation system: Bridging the gap between human and machine translation.
\newblock {\em arXiv preprint arXiv:1609.08144}, 2016.

\bibitem{iannantuono2023applications}
Giovanni~Maria Iannantuono, Dara Bracken-Clarke, Charalampos~S Floudas, Mario Roselli, James~L Gulley, and Fatima Karzai.
\newblock Applications of large language models in cancer care: current evidence and future perspectives.
\newblock {\em Frontiers in Oncology}, 13:1268915, 2023.

\bibitem{sinha2018drug}
Sandeep Sinha and Divya Vohora.
\newblock Drug discovery and development: An overview.
\newblock {\em Pharmaceutical Medicine and Translational Clinical Research}, pages 19--32, 2018.

\bibitem{pal2023chatgpt}
Soumen Pal, Manojit Bhattacharya, Md~Aminul Islam, and Chiranjib Chakraborty.
\newblock {ChatGPT} or {LLM} in next-generation drug discovery and development: pharmaceutical and biotechnology companies can make use of the artificial intelligence-based device for a faster way of drug discovery and development.
\newblock {\em International Journal of Surgery}, 109(12):4382--4384, 2023.

\bibitem{chakraborty2023artificial}
Chiranjib Chakraborty, Manojit Bhattacharya, and Sang-Soo Lee.
\newblock Artificial intelligence enabled {ChatGPT} and large language models in drug target discovery, drug discovery, and development.
\newblock {\em Molecular Therapy-Nucleic Acids}, 33:866--868, 2023.

\bibitem{stade2024large}
Elizabeth~C Stade, Shannon~Wiltsey Stirman, Lyle~H Ungar, Cody~L Boland, H~Andrew Schwartz, David~B Yaden, Jo{\~a}o Sedoc, Robert~J DeRubeis, Robb Willer, and Johannes~C Eichstaedt.
\newblock Large language models could change the future of behavioral healthcare: a proposal for responsible development and evaluation.
\newblock {\em NPJ Mental Health Research}, 3(1):12, 2024.

\bibitem{arslan2023exploring}
Sedat Arslan.
\newblock Exploring the potential of {ChatGPT} in personalized obesity treatment.
\newblock {\em Annals of Biomedical Engineering}, 51(9):1887--1888, 2023.

\bibitem{quer2024potential}
Giorgio Quer and Eric~J Topol.
\newblock The potential for large language models to transform cardiovascular medicine.
\newblock {\em The Lancet Digital Health}, 6(10):e767--e771, 2024.

\bibitem{benary2023leveraging}
Manuela Benary, Xing~David Wang, Max Schmidt, Dominik Soll, Georg Hilfenhaus, Mani Nassir, Christian Sigler, Maren Kn{\"o}dler, Ulrich Keller, Dieter Beule, et~al.
\newblock Leveraging large language models for decision support in personalized oncology.
\newblock {\em JAMA Network Open}, 6(11):e2343689--e2343689, 2023.

\bibitem{zeng2020meddialog}
Guangtao Zeng, Wenmian Yang, Zeqian Ju, Yue Yang, Sicheng Wang, Ruisi Zhang, Meng Zhou, Jiaqi Zeng, Xiangyu Dong, Ruoyu Zhang, et~al.
\newblock {MedDialog}: Large-scale medical dialogue datasets.
\newblock In {\em The Conference on Empirical Methods in Natural Language Processing}, pages 9241--9250, 2020.

\bibitem{rawte2023survey}
Vipula Rawte, Amit Sheth, and Amitava Das.
\newblock A survey of hallucination in large foundation models.
\newblock {\em arXiv preprint arXiv:2309.05922}, 2023.

\bibitem{amer2023large}
Sihem Amer-Yahia, Angela Bonifati, Lei Chen, Guoliang Li, Kyuseok Shim, Jianliang Xu, and Xiaochun Yang.
\newblock From large language models to databases and back: A discussion on research and education.
\newblock {\em ACM SIGMOD Record}, 52(3):49--56, 2023.

\bibitem{lin2023llm}
Yen-Ting Lin and Yun-Nung Chen.
\newblock {LLM-Eval}: Unified multi-dimensional automatic evaluation for open-domain conversations with large language models.
\newblock In {\em The 5th Workshop on NLP for Conversational AI}, pages 47--58, 2023.

\bibitem{sison2024chatgpt}
Alejo Jose~G Sison, Marco~Tulio Daza, Roberto Gozalo-Brizuela, and Eduardo~C Garrido-Merch{\'a}n.
\newblock {ChatGPT}: More than a “weapon of mass deception” ethical challenges and responses from the human-centered artificial intelligence {(HCAI)} perspective.
\newblock {\em International Journal of Human--Computer Interaction}, 40(17):4853--4872, 2024.

\bibitem{zhui2024ethical}
Zhui Li, Fenghe Li, Xuehu Wang, Qining Fu, and Wei Ren.
\newblock Ethical considerations and fundamental principles of large language models in medical education.
\newblock {\em Journal of Medical Internet Research}, 26:e60083, 2024.

\bibitem{el2011systematic}
Khaled El~Emam, Elizabeth Jonker, Luk Arbuckle, and Bradley Malin.
\newblock A systematic review of re-identification attacks on health data.
\newblock {\em PloS One}, 6(12):e28071, 2011.

\bibitem{ford2020should}
Elizabeth Ford, Malcolm Oswald, Lamiece Hassan, Kyle Bozentko, Goran Nenadic, and Jackie Cassell.
\newblock Should free-text data in electronic medical records be shared for research? a citizens’ jury study in the uk.
\newblock {\em Journal of Medical Ethics}, 46(6):367--377, 2020.

\bibitem{parviainen2022chatbot}
Jaana Parviainen and Juho Rantala.
\newblock Chatbot breakthrough in the 2020s? an ethical reflection on the trend of automated consultations in health care.
\newblock {\em Medicine, Health Care and Philosophy}, 25(1):61--71, 2022.

\bibitem{kerasidou2021need}
Angeliki Kerasidou, Kristine B{\ae}r{\o}e, Zackary Berger, and Amy E~Caruso Brown.
\newblock The need for empathetic healthcare systems.
\newblock {\em Journal of Medical Ethics}, 47(12):e27--e27, 2021.

\bibitem{harrer2023attention}
Stefan Harrer.
\newblock Attention is not all you need: the complicated case of ethically using large language models in healthcare and medicine.
\newblock {\em EBioMedicine}, 90, 2023.

\bibitem{reddy2023evaluating}
Sandeep Reddy.
\newblock Evaluating large language models for use in healthcare: A framework for translational value assessment.
\newblock {\em Informatics in Medicine Unlocked}, 41:101304, 2023.

\bibitem{munn2024truth}
Luke Munn, Liam Magee, and Vanicka Arora.
\newblock Truth machines: synthesizing veracity in {AI} language models.
\newblock {\em AI \& Society}, 39(6):2759--2773, 2024.

\bibitem{atallah2023large}
SB~Atallah, NR~Banda, A~Banda, and NA~Roeck.
\newblock How large language models including generative pre-trained transformer {(GPT)} 3 and 4 will impact medicine and surgery.
\newblock {\em Techniques in Coloproctology}, 27(8):609--614, 2023.

\bibitem{wang2019bio}
Lili Wang, Zheng Lou, Kai Jiang, and Guozhen Shen.
\newblock Bio-multifunctional smart wearable sensors for medical devices.
\newblock {\em Advanced Intelligent Systems}, 1(5):1900040, 2019.

\bibitem{huang2023intelligent}
Rong Huang, Hongxiu Li, Reima Suomi, Chenglong Li, and Teijo Peltoniemi.
\newblock Intelligent physical robots in health care: systematic literature review.
\newblock {\em Journal of Medical Internet Research}, 25:e39786, 2023.

\bibitem{preum2021review}
Sarah~Masud Preum, Sirajum Munir, Meiyi Ma, Mohammad~Samin Yasar, David~J Stone, Ronald Williams, Homa Alemzadeh, and John~A Stankovic.
\newblock A review of cognitive assistants for healthcare: Trends, prospects, and future directions.
\newblock {\em ACM Computing Surveys}, 53(6):1--37, 2021.

\bibitem{chen2022metaverse}
Zefeng Chen, Jiayang Wu, Wensheng Gan, and Zhenlian Qi.
\newblock Metaverse security and privacy: An overview.
\newblock In {\em International Conference on Big Data}, pages 2950--2959. IEEE, 2022.

\bibitem{chen2024metaverse}
Zefeng Chen, Wensheng Gan, Jiayang Wu, Hong Lin, and Chien-Ming Chen.
\newblock Metaverse for smart cities: A survey.
\newblock {\em Internet of Things and Cyber-Physical Systems}, 2024.

\bibitem{he2023large}
Jingxuan He and Martin Vechev.
\newblock Large language models for code: Security hardening and adversarial testing.
\newblock In {\em The ACM SIGSAC Conference on Computer and Communications Security}, pages 1865--1879, 2023.

\bibitem{roman2018blockchain}
Juan~M Roman-Belmonte, Hortensia De~la Corte-Rodriguez, and E~Carlos Rodriguez-Merchan.
\newblock How blockchain technology can change medicine.
\newblock {\em Postgraduate Medicine}, 130(4):420--427, 2018.

\bibitem{wu2023multimodal}
Jiayang Wu, Wensheng Gan, Zefeng Chen, Shicheng Wan, and Phillip~S Yu.
\newblock Multimodal large language models: A survey.
\newblock In {\em International Conference on Big Data}, pages 2247--2256. IEEE, 2023.

\bibitem{mendoza2024adaptive}
Rafael Mendoza, Isabella Cruz, Richard Liu, Aarav Deshmukh, David Williams, Jesscia Peng, and Rohan Iyer.
\newblock Adaptive self-supervised learning strategies for dynamic on-device {LLM} personalization.
\newblock {\em arXiv preprint arXiv:2409.16973}, 2024.

\bibitem{yin2024survey}
Shukang Yin, Chaoyou Fu, Sirui Zhao, Ke~Li, Xing Sun, Tong Xu, and Enhong Chen.
\newblock A survey on multimodal large language models.
\newblock {\em National Science Review}, 11(12):nwae403, 2024.

\bibitem{li2024llava}
Chunyuan Li, Cliff Wong, Sheng Zhang, Naoto Usuyama, Haotian Liu, Jianwei Yang, Tristan Naumann, Hoifung Poon, and Jianfeng Gao.
\newblock {LLaVA-Med}: Training a large language-and-vision assistant for biomedicine in one day.
\newblock {\em Advances in Neural Information Processing Systems}, 36, 2024.

\bibitem{shu2023visual}
Chang Shu, B~Chen, F~Liu, Z~Fu, E~Shareghi, and N~Collier.
\newblock Visual {Med-Alpaca}: A parameter-efficient biomedical {LLM} with visual capabilities, 2023.

\bibitem{moor2023med}
Michael Moor, Qian Huang, Shirley Wu, Michihiro Yasunaga, Yash Dalmia, Jure Leskovec, Cyril Zakka, Eduardo~Pontes Reis, and Pranav Rajpurkar.
\newblock {Med-Flamingo}: a multimodal medical few-shot learner.
\newblock In {\em Machine Learning for Health}, pages 353--367. PMLR, 2023.

\bibitem{liu2023qilin}
Junling Liu, Ziming Wang, Qichen Ye, Dading Chong, Peilin Zhou, and Yining Hua.
\newblock {Qilin-Med-VL}: Towards chinese large vision-language model for general healthcare.
\newblock {\em arXiv preprint arXiv:2310.17956}, 2023.

\bibitem{li2024frozen}
Jun Li, Che Liu, Sibo Cheng, Rossella Arcucci, and Shenda Hong.
\newblock Frozen language model helps {ECG} zero-shot learning.
\newblock In {\em Medical Imaging with Deep Learning}, pages 402--415. PMLR, 2024.

\bibitem{englhardt2024exploring}
Zachary Englhardt, Richard Li, Dilini Nissanka, Zhihan Zhang, Girish Narayanswamy, Joseph Breda, Xin Liu, Shwetak Patel, and Vikram Iyer.
\newblock Exploring and characterizing large language models for embedded system development and debugging.
\newblock In {\em Extended Abstracts of the CHI Conference on Human Factors in Computing Systems}, pages 1--9, 2024.

\bibitem{wang2025evaluation}
Xueqi Wang, Haiyan Ye, Sumian Zhang, Mei Yang, and Xuebin Wang.
\newblock Evaluation of the performance of three large language models in clinical decision support: A comparative study based on actual cases.
\newblock {\em Journal of Medical Systems}, 49(1):23, 2025.

\bibitem{chen2023extensive}
Qijie Chen, Haotong Sun, Haoyang Liu, Yinghui Jiang, Ting Ran, Xurui Jin, Xianglu Xiao, Zhimin Lin, Hongming Chen, and Zhangmin Niu.
\newblock An extensive benchmark study on biomedical text generation and mining with {ChatGPT}.
\newblock {\em Bioinformatics}, 39(9):btad557, 2023.

\bibitem{zhou2023survey}
Hongjian Zhou, Fenglin Liu, Boyang Gu, Xinyu Zou, Jinfa Huang, Jinge Wu, Yiru Li, Sam~S Chen, Peilin Zhou, Junling Liu, et~al.
\newblock A survey of large language models in medicine: Progress, application, and challenge.
\newblock {\em arXiv preprint arXiv:2311.05112}, 2023.

\bibitem{zhao2025technological}
Xinlei Zhao.
\newblock Technological hedging and differentiated responses of southeast asian countries to us--china technological competition: A case study on artificial intelligence (ai).
\newblock {\em The Pacific Review}, 38(3):502--533, 2025.

\end{thebibliography}

\end{document}